\documentclass[runningheads]{llncs}
\usepackage[utf8]{inputenc}
\usepackage{xcolor}
\usepackage{amsmath}\usepackage{amssymb}
\usepackage{url}
\usepackage{mdframed}
\usepackage{subcaption}
\usepackage{tikz,pgfplots}
\usepackage{paralist}
\usepackage{microtype}
\usepackage{nicefrac}
\usepackage{algorithm}
\usepackage{algpseudocode}
\usepackage{listings}
\usepackage{colonequals}
\usetikzlibrary{automata,positioning,shapes,fit,calc}
\usetikzlibrary{backgrounds}

\newcommand{\ntparti}{\mathbb{I}}
\newcommand{\Sched}{\Sigma}

\newcommand{\mdp}{\mathcal{M}}
\newcommand{\act}{\alpha}
\newcommand{\Act}{A}
\newcommand{\EnAct}{\textsf{Act}}
\renewcommand{\path}{\pi}
\newcommand{\pr}{Pr}
\newcommand{\last}[1]{\textsf{last}(#1)}
\newcommand{\pred}[2]{\textsf{pred}_#1(#2)}
\renewcommand{\succ}[2]{\textsf{succ}_{#1}(#2)}
\newcommand{\finpaths}[1]{\textsf{Paths}_\textsf{fin}}
\newcommand{\infpaths}[1]{\textsf{Paths}_\textsf{fin}}

\newcommand{\classmdp}{\mathcal{T}}
\newcommand{\exprew}[1]{\textsf{ER}_{#1}}
\newcommand{\prob}[1]{\textsf{Pr}_{#1}}
\newcommand{\maxexprew}[1]{\textsf{ER}_{#1}^\textsf{max}}
\newcommand{\sched}{\sigma}
\newcommand{\succs}[1]{\textsf{succ}(#1)}
\newcommand{\parsched}{\hat{\sigma}}
\newcommand{\macro}[1]{\mu(#1)}
\newcommand{\umacro}[1]{\nu(#1)}
\newcommand{\entry}{\textsf{entry}}
\newcommand{\lb}{\textsf{lb}}
\newcommand{\ub}{\textsf{ub}}
\newcommand{\LB}{u^{-}}
\newcommand{\UB}{u^{+}}
\newcommand{\res}{\textsf{res}}
\newcommand{\lbres}{\textsf{lbres}}
\newcommand{\ubres}{\textsf{ubres}}
\newcommand{\restoreg}{\textsf{Reg}}
\newcommand{\eqdef}{\colonequals}
\definecolor{problem}{HTML}{c0fa8b}

\definecolor{commentgreen}{RGB}{2,112,10}
\definecolor{eminence}{RGB}{108,48,130}
\definecolor{weborange}{RGB}{255,165,0}
\definecolor{frenchplum}{RGB}{129,20,83}

\newcommand{\toReg}{\textsf{toRegion}}

\newcommand{\RR}{\mathbb{R}}

\tikzset{tstate/.style={circle, draw=black, fill=white, inner sep=2pt}}
\tikzset{sstate/.style={circle, draw=black, fill=white, inner sep=1pt,font=\scriptsize}}
\tikzset{mstate/.style={rectangle, draw=black, fill=green!30, inner sep=1pt,font=\scriptsize, minimum height=4mm}}
\tikzset{tact/.style={rectangle, draw=black, fill=black, inner sep=1pt}}

\pgfplotsset{compat=1.15,
    every axis/.append style={
            font=\large,
            line width=1pt,
            tick style={line width=0.8pt}}}

\title{Abstraction-Refinement for Hierarchical~Probabilistic~Models}
\author{Sebastian Junges\inst{1} \and Matthijs T.\ J.\ Spaan\inst{2}}
\institute{Radboud University, Nijmegen, the Netherlands \and
Delft University of Technology, Delft, the Netherlands}
\titlerunning{Abstraction-Refinement for Hierarchical~Probabilistic~Models}
\authorrunning{S.\ Junges and M.\ T.\ J.\ Spaan}

\begin{document}

\maketitle
\begin{abstract}
Markov decision processes are a ubiquitous formalism for modelling systems with non-deterministic and probabilistic behavior. 
Verification of these models is subject to the famous state space explosion problem. 
We alleviate this problem by exploiting a hierarchical structure with repetitive parts. 
This structure not only occurs naturally in robotics, but also in probabilistic programs describing, e.g., network protocols. Such programs often repeatedly call a subroutine with similar behavior. In this paper, we focus on a local case, in which the subroutines have a limited effect on the overall system state.
The key ideas to accelerate analysis of such programs are (1) to treat the behavior of the subroutine as uncertain and only remove this uncertainty by a detailed analysis if needed, and (2)
to abstract similar subroutines into a parametric template, and then analyse this template. These two ideas are embedded into an abstraction-refinement loop that analyses hierarchical MDPs. 
A~prototypical implementation shows the efficacy of the approach. 
\end{abstract}

\section{Introduction}
Markov Decision Processes (MDPs) are \emph{the} model for sequential decision making under probabilistic uncertainty, and as such are 
central in modelling of randomized algorithms, distributed systems with lossy channels, or as the underlying formalism in reinforcement learning. 
A key question in the verification of MDPs is: \emph{What is the maximal probability that some error state is reached?}. In this question, one accounts for the probabilistic nature as well as the inherit (potentially adversarial) nondeterminism of the system. 
Various state-of-the-art probabilistic model checkers, such as Storm~\cite{DBLP:journals/corr/abs-2002-07080}, Prism~\cite{DBLP:conf/cav/KwiatkowskaNP11} and Modest~\cite{DBLP:conf/tacas/HartmannsH14}
implement a variety of methods that automatically compute such maximal probabilities. Most widespread are variations of value-iteration that iteratively apply a transition function to converge towards the requested probability. 

\medskip
\noindent\emph{Hierarchical structure.}
Despite various successes, the state space explosion remains a significant challenge to the model-based analysis of MDPs. To overcome this challenge, some approaches  exploit symmetries or the parallel composition of a system. Other approaches exploit that typically not all paths through a system are equally likely and thus aim to find the essential or critical subsystem. While we exploit related ideas ---a detailed comparison is given in the related work, cf.\ Sec.~\ref{sec:related}--- our approach is fundamentally different and instead exploits a \emph{hierarchical decomposition} natural in many system models. This decomposition is captured naturally by probabilistic programs (over discrete bounded variables) with non-nested subroutines, where some subroutines are called repeatedly with \emph{similar} arguments. Figure~\ref{fig:example1} shows an example on which we demonstrate our approach in Sec.~\ref{sec:overview}. More generally, we are interested in systems with an overall task that is achieved by a suitable combination of a limited number of sub-tasks. Such a setting occurs naturally, e.g.\ (i) in robotics, when multiple rooms in a floor need to be inspected, or (ii) in routing, when multiple packets need to be routed sequentially. 
The underlying problem structure is also exploited in hierarchical planning~\cite{DBLP:conf/nips/PrecupS97,DBLP:conf/uai/HauskrechtMKDB98,DBLP:conf/ijcai/BarryKL11}, where the goal is to find a good but not necessarily optimal policy (and induced value). \emph{We combine insights from hierarchical planning with an abstraction-refinement perspective and then construct an anytime algorithm with strict guarantees on the result.}

\begin{figure}[t]
\centering
 \begin{subfigure}{0.49\textwidth}
\begin{lstlisting}
  p = 0.5; time = 0; N=3;
  repeat N times {
     time += passToken(p);
     if flip(0.5) {p = 0.8p} 
     else         {p = 1.25p} 
  }; return time
\end{lstlisting}
\caption{Repeated invocation of \texttt{\textcolor{blue}{passToken}(p)}}
\label{fig:example1:main}
\end{subfigure}
 \begin{subfigure}{0.49\textwidth}
  \begin{lstlisting}
  passToken(p):
    t = 1;
    while (not flip(p)) {t++};
    t++;
    while (not flip(p)) {t++};
    return t
  \end{lstlisting}
 \caption{\texttt{\textcolor{blue}{passToken}(p)}: Pass succeed twice.}
 \label{fig:passTokenProg}
 \end{subfigure}
 \caption{Simplified example for sending a token over an unreliable channel.}
 \label{fig:example1}
\end{figure}

\medskip
\noindent\emph{Local model-based analysis.}
An adequate operational model for the model-based analysis of hierarchical systems is given by a \emph{hierarchical MDP}, where the state space of an hierarchical MDP can be partitioned into \emph{subMDPs}. Abstractly, one can represent a hierarchical MDP by the collection of subMDPs and a \emph{macro-level MDP}~\cite{DBLP:conf/uai/HauskrechtMKDB98} where the probabilities of outgoing transitions at a state are described by a corresponding subMDP, cf.~Sec.~\ref{sec:hmdps}. 
In this paper, we focus on a hierarchical MDPs where the policies that are optimal in (only) a subMDP are  optimal (partial) policies in the hierarchical MDP. More intuitively, if we solve a subtask optimally (w.r.t. a well-defined but fixed measure), then that way of solving the subtask is also globally optimal. The measure for optimality must be the same for every task. 
While this assumption is restrictive, it is satisfied on various interesting settings.  The assumption allows us to analyse subMDPs out-of-context, i.e., we can first analyse the subMDPs and then construct the correct macro-MDP, i.e., extract transition probabilities and rewards from the subMDP analysis. This approach already improves the maximal memory consumption and allows for additional speed-ups if the \emph{same} subMDP occurs multiple times.

\medskip
\noindent\emph{Epistemic uncertainty during computation.}
The key insight to accelerate the outlined approach further is to avoid analysing all subMDPs precisely, while still providing sound guarantees on the obtained results. 
Therefore, consider that even before analysing the subMDPs we can analyse an uncertain variant of the macro-level MDP where we do not yet know the associated transition probabilities and rewards but instead only know intervals.  
We may then do two things: First, we can identify the subMDPs which are most critical, i.e., where replacing the interval by a concrete value yields most benefits. Second, and more important, we can analyse a set of subMDPs and refine the associated uncertainties, i.e., tighten the associated intervals. 
To support the analysis of sets of subMDPs, we observe that often, these subMDPs  are slight variations. In this paper, we represent them as parameterised instances of a particular templates that we define using parametric MDPs (pMDPs). The result is an anytime computation of upper and lower bounds on the expected reward in the hierarchical MDP.

\medskip
\noindent\emph{Contributions.}
In a nutshell, we explicitly allow for \emph{uncertainty} during the solving process to speed up the analysis of hierarchical MDPs. 
Concretely, we contribute a scalable approach to solve  hierarchical MDPs with many different subMDPs, in particular when these subMDPs are similar, but not the same. The approach resembles an abstraction-refinement loop where we abstract the hierarchical MDP in two layers and then refine the analysis of the lower layer to get a refined representation of the complete MDP. In every step, we can provide absolute error bounds. Our approach interprets the different subMDPs as a form of uncertainty. The efficient analysis originates from  progress made in the analysis of uncertain (or parametric) MDPs, and brings that progress into a novel setting. The empirical evaluation with a prototype shows the efficacy of the approach.

\section{Overview}
\label{sec:overview}
We clarify the approach and its applicability with a  motivating example that drastically abstracts a token passing process where the channel quality varies~\cite{DBLP:conf/srds/DombrowskiJKG16}.
 
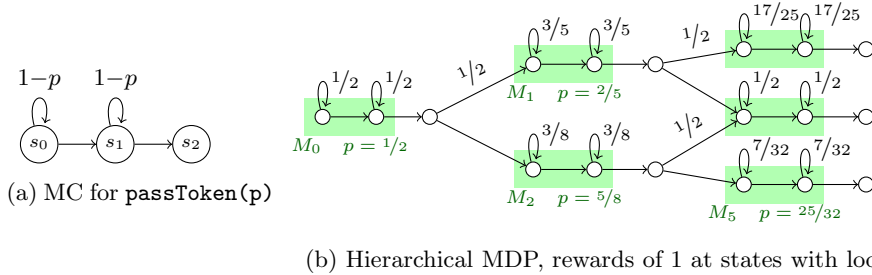
\begin{figure}[t]
\centering
\begin{subfigure}{0.29\textwidth}
\begin{tikzpicture}
	\node[tstate] at (0,0) (s0a) {\scriptsize{$s_0$}};
	\node[tstate,right=0.5cm of s0a] (s0b) {\scriptsize{$s_1$}};
	\node[tstate,right=0.5cm of s0b] (s0c) {\scriptsize{$s_2$}};
	
	\draw[->] (s0a) -- (s0b);
	\draw[->] (s0b) -- (s0c);
	\draw[->] (s0a) edge[loop above] node[above] {\footnotesize{$1{-}p$}}  (s0a);
	\draw[->] (s0b) edge[loop above] node[above] {\footnotesize{$1{-}p$}}  (s0b);

\end{tikzpicture}
\caption{MC for \texttt{passToken(p)}}
\label{fig:passToken}
\end{subfigure}
\begin{subfigure}{0.7\textwidth}
\begin{tikzpicture}[label distance=-3pt]
	\node[tstate] at (0,0) (s0a) {};
	\node[tstate,right=0.5cm of s0a] (s0b) {};
	\node[tstate,right=0.5cm of s0b] (s0c) {};
	
	\draw[->] (s0a) -- (s0b);
	\draw[->] (s0b) -- (s0c);
	\draw[->] (s0a) edge[loop above] node[right] {$\nicefrac{1}{2}$}  (s0a);
	\draw[->] (s0b) edge[loop above] node[right] {$\nicefrac{1}{2}$}  (s0b);
	
	\node[tstate] at (2.8,0.7) (s1a) {};
	\node[tstate,right=0.6cm of s1a] (s1b) {};
	\node[tstate,right=0.6cm of s1b] (s1c) {};
	
	\draw[->] (s1a) -- (s1b);
	\draw[->] (s1b) -- (s1c);
	\draw[->] (s1a) edge[loop above] node[right] {$\nicefrac{3}{5}$} (s1a);
	\draw[->] (s1b) edge[loop above] node[right] {$\nicefrac{3}{5}$} (s1b);
	
	\node[tstate] at (2.8,-0.7) (s2a) {};
	\node[tstate,right=0.6cm of s2a] (s2b) {};
	\node[tstate,right=0.6cm of s2b] (s2c) {};
	
	\draw[->] (s2a) -- (s2b);
	\draw[->] (s2b) -- (s2c);
	\draw[->] (s2a) edge[loop above] node[right] {$\nicefrac{3}{8}$} (s2a);
	\draw[->] (s2b) edge[loop above] node[right] {$\nicefrac{3}{8}$} (s2b);
	
	\node[tstate] at (5.6,0) (s3a) {};
	\node[tstate,right=0.6cm of s3a] (s3b) {};
	\node[tstate,right=0.6cm of s3b] (s3c) {};
	
	\draw[->] (s3a) -- (s3b);
	\draw[->] (s3b) -- (s3c);
	\draw[->] (s3a) edge[loop above] node[right] {$\nicefrac{1}{2}$} (s3a);
	\draw[->] (s3b) edge[loop above] node[right] {$\nicefrac{1}{2}$} (s3b);

	\node[tstate] at (5.6,0.9) (s4a) {};
	\node[tstate,right=0.6cm of s4a] (s4b) {};
	\node[tstate,right=0.6cm of s4b] (s4c) {};
	
	\draw[->] (s4a) -- (s4b);
	\draw[->] (s4b) -- (s4c);
	\draw[->] (s4a) edge[loop above] node[right] {$\nicefrac{17}{25}$} (s4a);
	\draw[->] (s4b) edge[loop above] node[right] {$\nicefrac{17}{25}$} (s4b);

	\node[tstate] at (5.6,-0.9) (s5a) {};
	\node[tstate,right=0.6cm of s5a] (s5b) {};
	\node[tstate,right=0.6cm of s5b] (s5c) {};
	
	\draw[->] (s5a) -- (s5b);
	\draw[->] (s5b) -- (s5c);
	\draw[->] (s5a) edge[loop above] node[right] {$\nicefrac{7}{32}$} (s5a);
	\draw[->] (s5b) edge[loop above] node[right] {$\nicefrac{7}{32}$} (s5b);

	\draw[->] (s0c) -- node[above,sloped] {$\nicefrac{1}{2}$} (s1a);
	
	\draw[->] (s0c) -- (s2a);
	\draw[->] (s1c) -- (s3a);
	\draw[->] (s1c) -- node[above,sloped] {$\nicefrac{1}{2}$} (s4a);
	\draw[->] (s2c) -- node[above,sloped] {$\nicefrac{1}{2}$} (s3a);
	\draw[->] (s2c) -- (s5a);
	
	\begin{scope}[on background layer]
	\node[fit=(s0a)(s0b),fill=green!30,inner sep=4pt,label={south:\raisebox{6pt}{\textcolor{green!40!black}{\scriptsize{$M_0\quad p=\nicefrac{1}{2}$}}}}] {};
	\node[fit=(s1a)(s1b),fill=green!30,inner sep=4pt,label={south:\textcolor{green!40!black}{\scriptsize{$M_1\quad p=\nicefrac{2}{5}$}}}] {};
		\node[fit=(s2a)(s2b),fill=green!30,inner sep=4pt,label={south:\textcolor{green!40!black}{\scriptsize{$M_2\quad p=\nicefrac{5}{8}$}}}] {};
		\node[fit=(s3a)(s3b),fill=green!30,inner sep=4pt] {};
		\node[fit=(s4a)(s4b),fill=green!30,inner sep=4pt] {};
		\node[fit=(s5a)(s5b),fill=green!30,inner sep=4pt,label={south:\textcolor{green!40!black}{\scriptsize{$M_5\quad p=\nicefrac{25}{32}$}}}] {};
	\end{scope}
\end{tikzpicture}
\caption{Hierarchical MDP, rewards of $1$ at states with loops}
\label{fig:hmdp}
\end{subfigure}	
\caption{Ingredients for hierarchical MDPs with the Example from Fig.~\ref{fig:example1}. Annotations reflect subMDPs within the macro-MDPs in Fig.~\ref{fig:hmdpanalysis}.}
\end{figure}

\medskip
\noindent\emph{Setting.}
Consider the protocol in Fig.~\ref{fig:example1:main} which sends a token $N$ times via a channel. That channel successfully transmits packets with probability $p$, where $p$ varies over time. The subroutine takes $t$ time, depending on $p$. 
Specifically, in the model, we alternate between accumulating the required time and updating the channel quality for $N$ token transmissions and then return the accumulated time.
We aim to compute the expected return value.
For the subroutine, we assume that sending a token is repeated until an acknowledgement is received, which is abstractly modelled in Fig.~\ref{fig:passTokenProg} and corresponds to the small Markov chain in Fig.~\ref{fig:passToken}. First, the file must successfully be sent ($s_0 \rightarrow s_1$), then we start sending acknowledgements. The process terminates ($s_1 \rightarrow s_2$) once an acknowledgement is received. 
The complete protocol from Fig.~\ref{fig:example1} including the subroutine is reflected by the large Markov chain in Fig.~\ref{fig:hmdp} that repeats the small Markov chain (with different probabilities). This model may be analysed with standard tools, but for large $N$ (and larger subroutines), the state space explosion must be alleviated.

\begin{figure}[t]
\centering
\begin{subfigure}{0.3\textwidth}
\scalebox{0.95}{
\begin{tikzpicture}
	\node[mstate,label=south:\scriptsize$4$, initial, initial text=, initial where=above] (m0) {$m_0$};
	\node[tstate,right=0.3cm of m0] (s0) {};
	\node[mstate,label=south:\scriptsize$5$] at (1.3,0.6) (m1) {$m_1$};
	\node[tstate,right=0.3cm of m1] (s1) {};
	\node[mstate,label=south:\scriptsize$\nicefrac{16}{5}$] at (1.3,-0.6) (m2) {$m_2$};
	\node[tstate,right=0.3cm of m2] (s2) {};
	\node[mstate,label=south:\scriptsize$\nicefrac{25}{4}$]  at (2.6,1) (m3) {$m_3$};
	\node[tstate,right=0.3cm of m3] (s3) {};
	\node[mstate,label=south:\scriptsize$4$] at (2.6,0)  (m4) {$m_4$};
	\node[tstate,right=0.3cm of m4] (s4) {};
	\node[mstate,label=south:\scriptsize$\nicefrac{64}{25}$] at (2.6,-1)  (m5) {$m_5$};
	\node[tstate,right=0.3cm of m5] (s5) {};
	
	\draw[->] (m0) -- (s0);
	\draw[->] (m1) -- (s1);
	\draw[->] (m2) -- (s2);
	\draw[->] (m3) -- (s3);
	\draw[->] (m4) -- (s4);
	\draw[->] (m5) -- (s5);
	\draw[->] (s0) -- (m1);
	\draw[->] (s0) -- (m2);
	\draw[->] (s1) -- (m3);
	\draw[->] (s1) -- (m4);
	\draw[->] (s2) -- (m4);
	\draw[->] (s2) -- (m5);
\end{tikzpicture}
}
\caption{}
\label{fig:hmdp:macro}
\end{subfigure}
\begin{subfigure}{0.34\textwidth}
\scalebox{0.95}{
\begin{tikzpicture}
	\node[mstate,label=south:\scriptsize${[\frac{64}{25},\frac{25}{4}]}$, initial, initial text=, initial where=above] (m0) {$m_0$};
	\node[tstate,right=0.3cm of m0] (s0) {};
	\node[mstate,label=south:\scriptsize${[\frac{64}{25},\frac{25}{4}]}$] at (1.3,0.6) (m1) {$m_1$};
	\node[tstate,right=0.3cm of m1] (s1) {};
	\node[mstate,label=south:\scriptsize${[\frac{64}{25},\frac{25}{4}]}$] at (1.3,-0.6) (m2) {$m_2$};
	\node[tstate,right=0.3cm of m2] (s2) {};
	\node[mstate,label=south:\scriptsize${[\frac{64}{25},\frac{25}{4}]}$]  at (2.6,1) (m3) {$m_3$};
	\node[tstate,right=0.3cm of m3] (s3) {};
	\node[mstate,label=south:\scriptsize${[\frac{64}{25},\frac{25}{4}]}$] at (2.6,0)  (m4) {$m_4$};
	\node[tstate,right=0.3cm of m4] (s4) {};
	\node[mstate,label=south:\scriptsize${[\frac{64}{25},\frac{25}{4}]}$] at (2.6,-1)  (m5) {$m_5$};
	\node[tstate,right=0.3cm of m5] (s5) {};
	
	\draw[->] (m0) -- (s0);
	\draw[->] (m1) -- (s1);
	\draw[->] (m2) -- (s2);
	\draw[->] (m3) -- (s3);
	\draw[->] (m4) -- (s4);
	\draw[->] (m5) -- (s5);
	\draw[->] (s0) -- (m1);
	\draw[->] (s0) -- (m2);
	\draw[->] (s1) -- (m3);
	\draw[->] (s1) -- (m4);
	\draw[->] (s2) -- (m4);
	\draw[->] (s2) -- (m5);
\end{tikzpicture}
}
\caption{}
\label{fig:hmdp:macroint}
\end{subfigure}
\begin{subfigure}{0.32\textwidth}
\scalebox{0.95}{
\begin{tikzpicture}
	\node[mstate,label=south:\scriptsize${[\frac{64}{25},4]}$, initial, initial text=, initial where=above] (m0) {$m_0$};
	\node[tstate,right=0.3cm of m0] (s0) {};
	\node[mstate,label=south:\scriptsize${[5,\frac{25}{4}]}$] at (1.3,0.6) (m1) {$m_1$};
	\node[tstate,right=0.3cm of m1] (s1) {};
	\node[mstate,label=south:\scriptsize${[\frac{64}{25},4]}$] at (1.3,-0.6) (m2) {$m_2$};
	\node[tstate,right=0.3cm of m2] (s2) {};
	\node[mstate,label=south:\scriptsize${[5,\frac{25}{4}]}$]  at (2.6,1) (m3) {$m_3$};
	\node[tstate,right=0.3cm of m3] (s3) {};
	\node[mstate,label=south:\scriptsize${[\frac{64}{25},4]}$] at (2.6,0)  (m4) {$m_4$};
	\node[tstate,right=0.3cm of m4] (s4) {};
	\node[mstate,label=south:\scriptsize${[\frac{64}{25},4]}$] at (2.6,-1)  (m5) {$m_5$};
	\node[tstate,right=0.3cm of m5] (s5) {};
	
	\draw[->] (m0) -- (s0);
	\draw[->] (m1) -- (s1);
	\draw[->] (m2) -- (s2);
	\draw[->] (m3) -- (s3);
	\draw[->] (m4) -- (s4);
	\draw[->] (m5) -- (s5);
	\draw[->] (s0) -- (m1);
	\draw[->] (s0) -- (m2);
	\draw[->] (s1) -- (m3);
	\draw[->] (s1) -- (m4);
	\draw[->] (s2) -- (m4);
	\draw[->] (s2) -- (m5);
\end{tikzpicture}
}
\caption{}
\label{fig:hmdp:macroref}
\end{subfigure}
\begin{subfigure}{0.34\textwidth}
\begin{tikzpicture}[xscale=3,yscale=0.2]
\draw[->] (0.13,0) -- (0.14,0) -- (0.16,-0.5)-- (0.18,0.5) -- (0.2,0) -- (1,0) node[right] {$p$};
\draw[->] (0.13,0) -- (0.13,12) node[left] {$\mathbb{E}$};
 \draw[domain=0.18:0.9, smooth, variable=\x, blue] plot ({\x}, {2/\x});
\draw[-,dotted] (0.32,0) node[below] {\rotatebox{270}{\tiny$\nicefrac{8}{25}$}} -- (0.32,6.25);
\draw[-,dotted] (0.4,0) node[below] {\rotatebox{270}{\tiny$\nicefrac{2}{5}$}} -- (0.4,5);
\draw[-,dotted] (0.5,0) node[below] {\rotatebox{270}{\tiny$\nicefrac{1}{2}$}} -- (0.5,4);
\draw[-,dotted] (0.625,0) node[below] {\rotatebox{270}{\tiny$\nicefrac{5}{8}$}} -- (0.625,3.2);
\draw[-,dotted] (0.78125,0) node[below] {\rotatebox{270}{\tiny$\nicefrac{25}{32}$}} -- (0.78125,2.56);

\draw[-,dotted] (0.32,6.25) -- (0.13,6.25)  node[left] {\rotatebox{0}{\tiny$\nicefrac{25}{4}$}};
\draw[-,dotted] (0.4,5)  -- (0.13,5) node[left] {\rotatebox{0}{\tiny$5$}};
\draw[-,dotted] (0.5,4)  -- (0.13,4) node[left] {\rotatebox{0}{\tiny$4$}};
\draw[-,dotted] (0.625,3.2) -- (0.13,3.2)node[left] {\rotatebox{0}{\tiny$\nicefrac{16}{5}$}};
\draw[-,dotted] (0.78125,2.56) -- (0.13,2.56) node[left] {\rotatebox{0}{\tiny$\nicefrac{64}{25}$}};

\end{tikzpicture}
\caption{}
\label{fig:hmdp:function}
\end{subfigure}
\begin{subfigure}{0.32\textwidth}
\begin{tikzpicture}[xscale=3,yscale=0.2]
\draw[->] (0.13,0) -- (0.14,0) -- (0.16,-0.5)-- (0.18,0.5) -- (0.2,0) -- (1,0) node[right] {$p$};
\draw[->] (0.13,0) -- (0.13,12) node[left] {$\mathbb{E}$};
 \draw[domain=0.18:0.9, smooth, variable=\x, blue] plot ({\x}, {2/\x});
\draw[-,dotted] (0.32,0) node[below] {\rotatebox{270}{\tiny$\nicefrac{8}{25}$}} -- (0.32,6.25);
\draw[-,dotted] (0.4,0) -- (0.4,5);
\draw[-,dotted] (0.5,0)  -- (0.5,4);
\draw[-,dotted] (0.625,0)  -- (0.625,3.2);
\draw[-,dotted] (0.78125,0) node[below] {\rotatebox{270}{\tiny$\nicefrac{25}{32}$}}  -- (0.78125,2.56);

\draw[-,dotted] (0.32,6.25) -- (0.13,6.25)  node[left] {\rotatebox{0}{\tiny$\nicefrac{25}{4}$}};
\draw[-,dotted] (0.78125,2.56) -- (0.13,2.56) node[left] {\rotatebox{0}{\tiny$\nicefrac{64}{25}$}};

 \fill [gray!50, domain=0.32:0.78124, variable=\x]
      (0.32, 0)
      -- plot ({\x}, {2/\x})
      -- (0.78125, 0)
      -- cycle;
  \fill[gray!50] (0.13,6.25)
  	  -- (0.32,6.25)
  	  -- (0.32,2.56)
  	  -- (0.13,2.56)
  	  -- cycle;

\end{tikzpicture}
\caption{}
\label{fig:hmdp:functionall}
\end{subfigure}
\begin{subfigure}{0.32\textwidth}
\begin{tikzpicture}[xscale=3,yscale=0.2]
\draw[->] (0.13,0) -- (0.14,0) -- (0.16,-0.5)-- (0.18,0.5) -- (0.2,0) -- (1,0) node[right] {$p$};
\draw[->] (0.13,0) -- (0.13,12) node[left] {$\mathbb{E}$};
 \draw[domain=0.18:0.9, smooth, variable=\x, blue] plot ({\x}, {2/\x});
\draw[-,dotted] (0.32,0) node[below] {\rotatebox{270}{\tiny$\nicefrac{8}{25}$}} -- (0.32,6.25);
\draw[-,dotted] (0.4,0) node[below] {\rotatebox{270}{\tiny$\nicefrac{2}{5}$}}  -- (0.4,5);
\draw[-,dotted] (0.5,0) node[below] {\rotatebox{270}{\tiny$\nicefrac{1}{2}$}} -- (0.5,4);
\draw[-,dotted] (0.625,0) -- (0.625,3.2);
\draw[-,dotted] (0.78125,0) node[below] {\rotatebox{270}{\tiny$\nicefrac{25}{32}$}}  -- (0.78125,2.56);

\draw[-,dotted] (0.32,6.25) -- (0.13,6.25)  node[left] {\rotatebox{0}{\tiny$\nicefrac{25}{4}$}};
\draw[-,dotted] (0.78125,2.56) -- (0.13,2.56) node[left] {\rotatebox{0}{\tiny$\nicefrac{64}{25}$}};

\draw[-,dotted] (0.32,6.25) -- (0.13,6.25)  node[left] {\rotatebox{0}{\tiny$\nicefrac{25}{4}$}};
\draw[-,dotted] (0.4,5) -- (0.13,5) node[left] {\rotatebox{0}{\tiny$5$}};
\draw[-,dotted] (0.5,4) -- (0.13,4) node[left] {\rotatebox{0}{\tiny$4$}};
\draw[-,dotted] (0.78125,2.56) -- (0.13,2.56) node[left] {\rotatebox{0}{\tiny$\nicefrac{64}{25}$}};

 \fill [gray!50, domain=0.5:0.78124, variable=\x]
      (0.5, 0)
      -- plot ({\x}, {2/\x})
      -- (0.78125, 0)
      -- cycle;
  \fill[gray!50] (0.13,4)
  	  -- (0.5,4)
  	  -- (0.5,2.56)
  	  -- (0.13,2.56)
  	  -- cycle;

 \fill [gray!50, domain=0.32:0.4, variable=\x]
      (0.32, 0)
      -- plot ({\x}, {2/\x})
      -- (0.4, 0)
      -- cycle;
  \fill[gray!50] (0.13,6.25)
  	  -- (0.32,6.25)
  	  -- (0.32,5)
  	  -- (0.13,5)
  	  -- cycle;

\end{tikzpicture}
\caption{}
\label{fig:hmdp:functionref}
\end{subfigure}
\caption{Visualising the computation of expected rewards for the hMDP from Fig.~\ref{fig:hmdp} using a macro-MDP and interval-based abstractions.}
\label{fig:hmdpanalysis}
\end{figure}
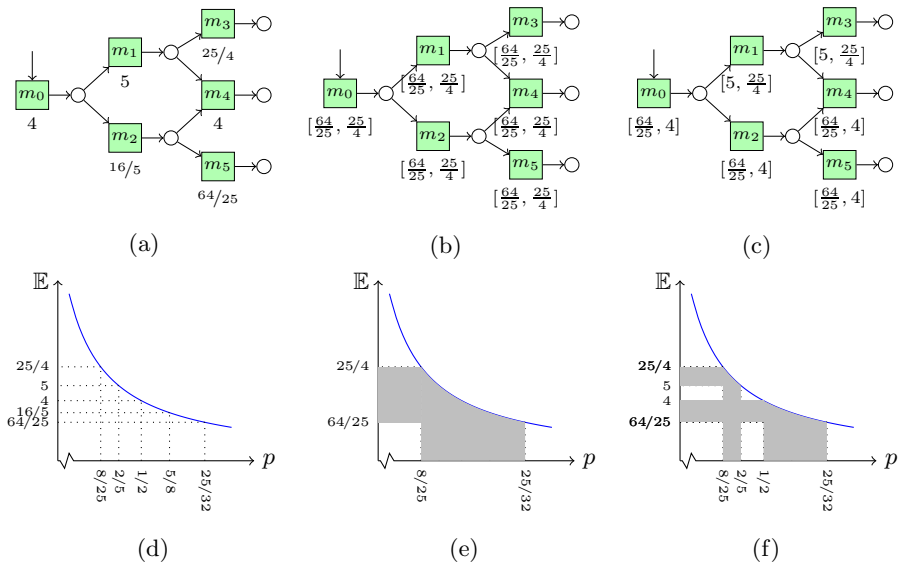

\medskip
\noindent\emph{Macro-MDPs and enumeration.}
We thus suggest to abstract the hierarchical model into the macro-level MDP in Fig.~\ref{fig:hmdp:macro}. Here, every state corresponds to an invocation of the subprocess. The reward at the states corresponds to the expected reward for the complete subprocess. Thus, naively, one may construct the macro-MDP, analyse all (reachable) subMDPs independently and annotate the macro-MDP states with the appropriate rewards, and finally analyse the macro-MDP to obtain a result of $\approx 12.3$. This approach avoids representing the complete hMDP in the memory, but it is still restricted to analysing systems with a limited number of subMDPs. 

\medskip
\noindent\emph{Our approach.}
We improve scalability by constructing a parameterized macro-MDP. Reconsider the rewards for Fig.~\ref{fig:hmdp:macro}. The values can be computed via the graph in Fig.~\ref{fig:hmdp:function}, where we pick for each value for $p$ (x-axis) and compute the corresponding expected reward $\mathbb{E}$ (y-axis) obtained by analysing the subMDP in  Fig.~\ref{fig:passToken}. 
Intuitively, in our abstraction,  we annotate the rewards with lower- and upper bounds rather than exact values. 
Therefore, we compute bounds on the rewards by selecting an interval for the values $p \in [\nicefrac{8}{25}, \nicefrac{25}{32}]$, as shown in Fig.~\ref{fig:hmdp:functionall}. Conceptually, this means that we analyse a set of subMDPs at once, namely all subMDPs with $p \in [\nicefrac{8}{25}, \nicefrac{25}{32}]$.
Annotating the corresponding expected rewards, in this case $[\nicefrac{64}{25}, \nicefrac{25}{4}]$, then yields the macro-MDP in Fig.~\ref{fig:hmdp:macroint}. Analysis of this MDP yields that overall expected time is in $[7.68, 18.75]$. 
We refine these bounds by analysing subsets of the subMDPs. We may split the values for $p$ into two sets $[\nicefrac{8}{25}, \nicefrac{2}{5}]$ and  $[\nicefrac{1}{2}, \nicefrac{25}{32}]$. Then, we obtain two corresponding intervals on the epxected time in the subMDP as shown in Fig.~\ref{fig:hmdp:functionref}. Model checking the associated macro-MDP, in Fig.~\ref{fig:hmdp:macroref}, bounds to expected time by $[10.12, 14.25]$.
Technically, we realize this reasoning using parameter lifting~\cite{DBLP:conf/atva/QuatmannD0JK16}.

%

    
\medskip
\noindent\emph{Supported extensions.}
For conciseness, this example is necessarily simple. Our approach allows  nondeterminism, i.e., action-choices, in the macro-MDP \emph{and} in the subMDPs. The subMDPs may have multiple outgoing transitions, but this must be combined with a restricted type of nondeterminism in the subMDP: If multiple outgoing transitions are present, the macro-MDP has transition probabilities that depend on the subMDPs. We present a useful extension for reachability probabilities, see the discussion at the bottom of Sec.~\ref{sec:localcase}.

\medskip
\noindent\emph{More examples.}
Key ingredient to models where the approach excels are a repetitive task whose characteristics depend on some global state. Two variations are the expected energy consumption of a robot with slowly degrading components that, e.g., can be improved by maintenance or for job scheduling with periodically changing distribution of tasks (e.g., day vs. night).

\section{Formal Problem Statement}
We formalize MDPs and \emph{hierarchical MDPs} (hMDPs) to pose the problem statement, then identify a subclass of hMDPs which we call \emph{local-policy hMDPs} and restrict our problem on computing optimal expected rewards in local-policy hMDPs. Furthermore, we introduce parametric MDPs as they are key to the abstraction-refinement procedure later in the paper. 

\subsection{Background}
\begin{definition}[Parametric MDP]
    A parametric MDP (pMDP) is a tuple $\mathcal{M} = \langle S_\mathcal{M}, \Act_\mathcal{M}, \iota_\mathcal{M}, \vec{x}, P_\mathcal{M}, r_\mathcal{M}, T_\mdp \rangle$ where $S_\mathcal{M}$ is a finite set of \emph{states}, $A_\mathcal{M}$ is a finite set of \emph{actions},  $\iota_\mathcal{M} \in S_\mathcal{S}$ is the \emph{initial state}, $\vec{x} = \langle x_0, \hdots x_n \rangle$ is a vector of \emph{parameters}, $P_\mathcal{M}\colon S_\mathcal{M} \times \Act_\mathcal{M} \times S_\mathcal{M} \rightarrow \mathbb{Q}[\vec{x}]$ are the \emph{transition probabilities},  $r_\mathcal{M}\colon S \rightarrow \mathbb{Q}[\vec{x}]$ the \emph{state rewards}, and $T_\mathcal{M}$ is a set of \emph{target states}.
\end{definition}
We drop the subscripts whenever possible. 
MDPs are \emph{parametric} if $\vec{x} \neq \langle \rangle$ and \emph{parameter-free} otherwise. We omit parameters for parameter-free MDPs. We recap some standard notions on pMDPs (and MDPs):

For an (parameter) \emph{valuation} $u \in \mathbb{R}^{\vec{x}}$, the \emph{instantiation} $\mathcal{M}[u]$ globally substitutes $P_\mathcal{M}(s,a,s')$ with $P_\mathcal{M}(s,a,s')(u)$ and $r_\mathcal{M}(s)$ with $r_\mathcal{M}(s)(u)$. 
An assignment $u$ is well-defined, if $\mathcal{M}(u)$ constitutes an MDP, i.e., if $\sum_{s'} P_\mathcal{\mdp}(s,\act,s')(u) \in \{0, 1\}$ and  $r_\mathcal{M}(s)(u) \geq 0$ for each $s \in S$, $\act \in \Act$.
We denote the set of all well-defined assignments with $U_\mdp$. The set $\EnAct(s)$ denotes the  enabled actions at state $s$, $\EnAct(s) = \{ \act \mid \sum_{s'} P_\mathcal{\mdp}(s,\act,s') \neq 0$ \}. If $|\EnAct(s)| = 1$ for every $s \in S$, then the (parametric) MDP is a (parametric) \emph{Markov chain} (MC).
 A path $\path$ is an (in)finite sequence of states $s_0 \xrightarrow{\act_0} s_1\hdots$, with $s_i \in S$, $\act_i \in \EnAct(s_i)$, $P(s_i, \act_i,s_{i+1}) \neq 0$.
For finite $\path$, $\last{\pi}$ denotes the last state of $\pi$. 
We use $[s \rightarrow \lozenge T]$ to denote the set of (finite) paths $T$ only at the end.
The reward $r(\pi)$ along a finite path  $\pi$ is the sum of the state rewards $r(\pi) \colonequals \sum r(s_i)$.

\paragraph{Specifications.}
We consider indefinite horizon expected reward, i.e.,, the expected accumulated reward until reaching the target states. 
We refer to \cite{Put95,BK08} for a formal treatment and only introduce notation. 
Therefore, the unique probability measure $\pr$ for a set of  paths in a parameter-free \emph{Markov chain} $\mdp$ reaching state $T$ can be defined using the usual cylinder set construction.
We define $\pr_\mdp(s \rightarrow \lozenge T)$ as the probability to reach a state in $T$, $\int_{\pi \in [s \rightarrow \lozenge T]} \pr(\pi)$.
We then define the expected reward until hitting $T$,  $\exprew{\mdp}(s \rightarrow \lozenge T) = \int_{\pi \in [s \rightarrow \lozenge T]} \pr(\pi) \cdot r(\pi)$. In both definitions, if $s$ is the initial state, we simply write $\hdots(\lozenge T)$.
For technical conciseness, we make the standard assumption that target states are reached with probability $1$, which ensures that the integral exists and is finite. (Arbitrary) reachability probabilities can be nevertheless be modelled using rewards.

\paragraph{Policies.}
In pMDPs, we resolve nondeterminism with policies. 
In this paper, it suffices to consider \emph{memoryless policies} $\sched\colon S \rightarrow \Act$. The set of such policies is denoted $\Sched(\mdp)$. We omit $\mdp$ if it is clear from the context. It is helpful to also consider \emph{partial} policies $\parsched\colon S \nrightarrow \Act$.
For an pMDP $\mdp$ and a (partial) policy $\parsched$, the induced dynamics are described by the \emph{induced pMDP} $\mdp[\parsched]$, defined as $\langle S_\mathcal{M}, \Act_\mathcal{M}, \iota_\mathcal{M}, \vec{x}, P, r_\mathcal{M}, T_\mdp \rangle$, where the transition probabilities are given as \[ P(s,\act, s') = \begin{cases} P_\mdp(s,\act,s')  & \text{if } \parsched(s) = \act, \\ 0 & \text{otherwise.} \end{cases} \]
If $\sched$ is total (not partial), then $\mdp$ is a MC.
We define the maximal expected reward  $\exprew{\mdp}^\text{max}(\lozenge T) = \max_{\sched \in \Sched}  \exprew{\mdp[\sched]}(\lozenge T)$, and say that  a policy $\sched$ is optimal, if $\exprew{\mdp}^\text{max}(\lozenge T) = \exprew{\mdp[\sched]}(\lozenge T)$.

\paragraph{Regions and parametric model checking.}
A set of valuations described by is called a (rectangular) \emph{region}, if $R = \{ u \mid \LB \leq u \leq \UB  \}$ for adequate bounds $\LB, \UB \in \RR^{\vec{x}}$ and using pointwise inequalities, 
i.e.,  $R$ is a Cartesian product of intervals of parameter values.
We denote this region also with $[[\LB,\UB]]$.
For regions, we may compute a lower bound on $\min_{u \in R} \maxexprew{\mdp[u]}(\lozenge T)$ and an upper bound on $\max_{u \in R} \maxexprew{\mdp[u]}(\lozenge T)$ via \emph{parameter lifting}~\cite{DBLP:conf/atva/QuatmannD0JK16,DBLP:conf/tacas/SpelJK21}.

\subsection{Hierarchical MDPs}
\label{sec:hmdps}
We concentrate on solving hierarchical MDPs (hMDPs). We assume that hMDPs are parameter-free and that their topology has some additional known structure.
\begin{definition}[Hierarchical MDPs]
\label{def:hmdp}
A MDP $\mdp$ with a partitioning of its states $S_\mdp = \bigcup \mathbf{S}_i$ is a hierarchical MDP, if for all $i$, 
\begin{compactitem}
    \item \text{there exists a unique $s^i_\iota \in \mathbf{S}_i$ such that   } $s^i_\iota = \iota_\mdp \text{ or } \pred{\mdp}{s^i_\iota} \not\subseteq \mathbf{S}_i$, and
    \item  $\text{for all } s \in \mathbf{S}_i \setminus \{ s^i_\iota \}$, it holds that $s^i_\iota \neq \iota_\mdp \text{ and } \pred{\mdp}{s} \subseteq \mathbf{S}_i.$
\end{compactitem}
\end{definition}
The state $s_\iota$ is called the \emph{entry state}, which we denote $\entry_i$. States with $\succ{\mdp}{s} \cap \mathbf{S}_i = \emptyset$ are called \emph{exit-states}. The set $\succs{i} \colonequals \succ{\mdp}{\mathbf{S}_i} \setminus \mathbf{S}_i$ are the \emph{successor states} of the partition $i$. Let $Y = \max_i |\succs{i}|$. By adding auxiliary states, we can assume that $|\succs{i}| = Y$ for all $i$.
We call partitions with $|\mathbf{S}_i| = 1$ \emph{trivial}. We use $\ntparti \colonequals \{ i \mid |\mathbf{S}_i| > 1 \}$ to denote the indices of the nontrivial partitions. 
We remark that every MDP can be considered as an hMDP with only trivial partitions.  
\begin{mdframed}[backgroundcolor=problem!10!white]
\textbf{Problem:} Given a (hierarchical) MDP $\mdp$ with target states $T$ and $\eta \in [0,1]$, compute bounds $\lb, \ub$ with $\lb \leq \maxexprew{\mdp}(\lozenge T) \leq \ub$ and $\eta \cdot \ub \leq \lb$. 
\end{mdframed}
The naive solution to this problem is to ignore the hierarchical structure and solve the MDP monolithically. 
In this paper, we contribute methods that actively exploit the structure of the hierarchical MDPs with $|\ntparti| \gg 1$. We will make an additional assumption on the structure of the hierarchical MDP.

\subsection{Optimal Local Subpolicies and Beyond}
\label{sec:localcase}
Intuitively, we want to ensure that the optimal policy within the partitions can be computed locally, i.e., on partition without taking into account the complete MDP. 
Therefore, each partition within the MDP can be considered as an individual MDP.
In particular, each $\mathbf{S}_i$ induces a subMDP as follows:
\begin{definition}[subMDP]
\label{def:submdp}
    Given a hierarchical MDP $\mdp$ and partition $\textbf{S}_i$, the corresponding subMDP is an MDP $\mdp_i \colonequals \langle S_i \colonequals \textbf{S}_i \cup \succ{\mdp}{\textbf{S}_i} \cup \{ \bot \}, \Act_\mdp \cup \{ \act_\bot \}, \iota \colonequals \entry_{i}, P_i, r_i, G_i \rangle$ with $P_i$ defined by
    \begin{align*}
        P_i(s,\act,s') \colonequals \begin{cases} 
        P_\mdp(s,\act, s') & \text{ if } s \in \textbf{S}_i \text{ and }\act \in \Act_\mdp, \\ 
        1 & \text{else if }  s \not\in \textbf{S}_i, 
       \act = \act_\bot, \text{ and } s' = \bot \\
      0  & \text{otherwise.}
        \end{cases}
        \end{align*}
        $r_i$ is defined as $r_i(s) = r_\mdp(s)$ if $s\in \textbf{S}_i$,  $r_i(s) = 0$ otherwise, and $G_i \colonequals \{ \bot_i \}$.
\end{definition}
Thus, for every partition of the hierarchical MDP, the corresponding subMDP contains additionally the successor states, and a unique bottom state that is a target state and simplifies our construction later.

Likewise, we can (de)compose memoryless policies for the hierarchical MDP as a union of policies on the individual subMDPs. We do this only for nontrivial partitions. Let $\sched_i \colon S_i \mapsto \Act$ denote memoryless policies for $\mdp_i$ and $\sched'_i$ the restriction of $\sched_i$ to $\textbf{S}_i$, then $\left( \bigsqcup_{\ntparti} \sched_{i } \right) \colon S \nrightarrow \Act$ is the unique partial policy such that \[ \big( \bigsqcup_{\ntparti} \sched_{i } \big)(s) \colonequals \sched'_i(s) \text{ if } s \in \textbf{S}_i,  i \in \ntparti \quad\text{ and }\quad \big( \bigsqcup_{\ntparti} \sched_{i } \big)(s) \colonequals \bot \text{ otherwise. } \]
Intuitively, we want that the union of locally optimal policies, a partial policy, can be completed to a total policy that is optimal. 
\begin{definition}[Optimal local subpolicies]
\label{def:optimal}
Given a hierarchical MDP $\mdp$ with target states $T$ and optimal policies $\sched_i \in \Sched(\mdp_i)$  for all $i \in \ntparti$.
The hierarchical MDP has \emph{optimal local subpolicies}, if for $\parsched = \bigsqcup_\ntparti \sched_i$ it holds that 
$\maxexprew{\mdp[\parsched]} = \maxexprew{\mdp}$.
\end{definition}
That is, if we collect (locally) optimal policies $\sched_i$ and apply them to $\mdp$, we obtain the MDP $\mdp[\left( \bigsqcup_{\ntparti} \sched_{i } \right)]$. In that MDP, we can pick an optimal policy, and together with $\left( \bigsqcup_{\ntparti} \sched_{i } \right)$ this constitutes an optimal and total policy for $\mdp$. 
\begin{mdframed}[backgroundcolor=problem!10!white]
\textbf{Assumption:} The hierarchical MDP has optimal local subpolicies.
\end{mdframed}
 Roughly, the idea now becomes that rather than solving one large MDP with $S$ states, we solve $|\ntparti|$ MDPs with $\nicefrac{S}{|\ntparti|}$ states and one MDP with $\ntparti$ states (assuming equally-sized and only nontrivial partitions). 
 
The assumption is restrictive, but not unreasonable: A subroutine may not have any nondeterminism, or a finished task will have no influence on any future task. The following proposition, while obvious, formalizes that:
\begin{proposition}[Sufficient criterion]
\label{prop:sufficient}
Let $\mdp$ be a hierarchical MDP. The MDP has optimal local subpolicies, if for each $i \in \ntparti$ either
\begin{compactitem}
    \item there is a single successor for the partition, i.e., $|\succ{\mdp}{\textbf{S}_i} \setminus \textbf{S}_i|=1$, or
    \item there are no choices, i.e., $|\EnAct(s)| = 1$ for all $s \in \textbf{S}_i$, 
\end{compactitem}
\end{proposition}

 \subsubsection{Beyond optimal local subpolicies.}
The efficiency of our approach is partly due to the assumption in Def.~\ref{def:optimal}. We observe that adapting this definition allows for a spectrum of specific yet useful cases. 
In particular, say that our system describes a protocol in which we must optimize the probability to satisfy $N$ tasks all may fail -- the subMDPs will have two successor states. Often, it is then easy to see (and model) that a locally optimal policy will aim to satisfy each task and that thus, the locally optimal policy optimizes the probability to reach the corresponding successor state. Then, by adopting the target states in Def.~\ref{def:submdp} to be the successor state where the task is successful, the notion of an optimal policy ---and thus of an optimal local subpolicy--- changes. These changes are minimal and everything that follows below is easily adapted to this setting as demonstrated by the prototypical implementation.

\section{Solving hMDPs with Abstraction-Refinement}
In this section, we consider hMDPs with optimal local subpolicies.  We step-wise develop a sketch of an anytime algorithm that provides lower and upper bounds on the expected reward in this hMDP. 
In Sec.~\ref{sec:macromdp}, we introduce an alternative representation of our problem that formalizes the idea of individually computing subMDPs. We then formalize the ideas that allow to construct an anytime algorithm in Sec.~\ref{sec:umacromdp}. Section~\ref{sec:setbasedanalysis} introduces the abstract requirements for analysing sets of subMDPs into the algorithm, whereas Sec.~\ref{sec:psubmdps} introduces a method that realises this using pMDPs.

 
\subsection{The Macro-MDP Formulation}
\label{sec:macromdp}
We adapt macro-MDPs~\cite{DBLP:conf/ijcai/BarryKL11} which summarize the subMDPs by single states. 
\begin{definition}[Macro-MDP]
\label{def:mmdp}
Let $\mdp$ be a hMDP with $n$ non-trivial $\textbf{S}_i$ partitions and $S_\mdp$ partitioned as $S_\mdp = \bigcup \textbf{S}_i \cup S'$.  The \emph{macro-MDP} is defined as 
$\macro{\mdp} \colonequals \langle S' \cup  \{ \entry_i \mid 1 \leq i \leq n \}, \Act_\mdp, \iota_\mdp, \emptyset, P, r, T_\mdp \rangle$ with $P$ and $r$ given by
\[ 
    P(s, \act, s') = \begin{cases} \prob{\mdp_i[\sigma_i]}(\lozenge \{s'\}) &\text{if } s \in \textbf{S}_i, \\ 
    P_\mdp(s,\act,s') & \text{otherwise,} \end{cases} \quad r(s) = \begin{cases} \maxexprew{\mdp_i}(\lozenge \{\bot \}) &\text{if } s \in \textbf{S}_i, \\ 
    r_\mdp(s) & \text{otherwise.}  \end{cases}
\]
where $\mdp_i$ is the corresponding subMDP (see Def.~\ref{def:submdp}) and $\sched_i$ is an arbitrary but fixed optimal policy, i.e, a policy such that $\exprew{\mdp_i[\sigma_i]}(\lozenge G_i) = \maxexprew{\mdp_i}(\lozenge G_i)$.
\end{definition}
Intuitively, we replace the transitions within $\textbf{S}_i$ by a `big-step semantics' that aggregates the transitions within $\textbf{S}_i$ by single transitions such that the probability to reach any successor matches the probability to do so within $\textbf{S}_i$ under a specific --optimal-- policy. Likewise, the expected reward matches the expected reward collected in $\textbf{S}_i$\footnote{Due to the additive nature of expected rewards, we can annotate the state with the expected reward even though it may differ over the different paths to an exit of $\textbf{S}_i$.}.


\begin{remark}
To define a \emph{unique} macro-MDP, we can take the lexicographically smallest policy $\sched_i$ among the optimal policies. Furthermore, we observe that for the cases covered by Prop.~\ref{prop:sufficient}, it is not necessary to compute $\sigma_i$ at all: Either there is a single successor ---implying $\prob{\mdp_i[\sigma_i]}(\lozenge \{s'\}) = 1$ for any  $\sched_i$--- or $|\Sched(\mdp_i)|=1$.
\end{remark}
The following theorem formalises that, given the assumptions, taking the big-step semantics is adequate when optimizing for an expected reward. 
\begin{theorem}
\label{thm:macrocorrect}
Let $\mdp$ be a hMDP with optimal local subpolicies and let $\macro{\mdp}$ be the corresponding macro-MDP. Then: $\maxexprew{\macro{\mdp}}(\lozenge T) = \maxexprew{\mdp}(\lozenge T)$.
\end{theorem}
The important ingredient are the optimal local subpolicies that ensure that we aggregate behavior within the partitions by behavior that agrees with a (globally) optimal policy.
We give a proof in App.~\ref{app:thmmacrocorrect}.

\paragraph{Naive Algorithm.}
Algorithmically, we first compute $\maxexprew{\mdp_i}(\lozenge T_i)$ and the associated policy $\sched_i$, then compute the reachability probabilities on the induced Markov chain. We collect these results in a a vector $\res_i$, which is helpful to construct the macro-MDP. To clarify further constructions in this paper,  we make $\res_i$ explicit.
Recall that $|\succ{\mdp}{\textbf{S}_i}| = Y$ for all $i$. 
\begin{definition}[Results for subMDP]
\label{def:results}
    Let $\mdp_i$ be a subMDP for the partition~$\textbf{S}_i$ of a hMDP $\mdp$. Let $\succ{\mdp}{\textbf{S}_i}$ be ordered. We define $\res_i \in \mathbb{R}^{Y+1}$ s.t.\ \[ \res_i(j) \colonequals  \prob{\mdp_i[\sigma_i]}(\lozenge \{\succ{\mdp}{\textbf{S}_i}_j \}) \text{ for }0 \leq  j < Y \text{ and } \res_i(Y) \colonequals \maxexprew{\mdp_i}(\lozenge G_i),\] where $\sched_i$ is an arbitrary but fixed policy such that $\exprew{\mdp_i[\sigma_i]}(\lozenge G_i) = \maxexprew{\mdp_i}(\lozenge G_i)$.
\end{definition}
This allows us to reformulate the macro-MDP, in particular, the following two identities do hold: 
\begin{align}
\label{eq:mmdpwresults}
    P(s, \act, s') {=} \begin{cases} \res_i(j) &\text{if } s \in \textbf{S}_i \text{ and } \\ & ~s' = \succ{\mdp}{\textbf{S}_i}_j \\ 
    P_\mdp(s,\act,s') & \text{otherwise,} \end{cases}
    \quad r(s) {=} \begin{cases} \res_i(Y) &\text{if } s \in \textbf{S}_i, \\ 
    r_\mdp(s) & \text{otherwise.}  \end{cases}
\end{align}
The identities trivialize that constructing the macro-MDP can be done by precomputing the necessary result-vectors.

\begin{mdframed}
\textbf{Enumeration baseline}: With macro-MDPs, we reduce the computation of $\maxexprew{\mdp}(\lozenge T)$ to (1) analysing all subMDPs $\mdp_i$ and (2) analysing $\macro{\mdp}$.
\end{mdframed}

This rather naive algorithm already limits memory and may exploit similarities between subMDPs during the analysis, e.g., based on the structure discussed in Sec.~\ref{sec:psubmdps}. It performs well if the number $|\ntparti|$ of subMDPs is sufficiently small. We are interested in considering methods that allow for larger $\ntparti$ or larger subMDPs. 
In particular, we want to avoid analysing all subMDPs, all individually. 

\subsection{The uncertain Macro-MDP Formulation}
\paragraph{Uncertainty before computation.}
\label{sec:umacromdp}

We start introducing a method that allows providing bounds on the expected rewards after individually analysing a subset of the subMDPs.
Before computing the individual probabilities in $\mdp_i$, we are \emph{uncertain} about the probabilities and rewards in the MDP $\macro{\mdp}$. Under this uncertainty, we may not be able to compute $\maxexprew{\macro{\mdp}}(\lozenge T)$ precisely. However, we may solve the problem statement by \emph{bounding} the expected reward. Thus, the goal is to compute values $\lb, \ub$ s.t.\  
\begin{align}
    \lb \leq  \maxexprew{\mdp}(\lozenge T) = \maxexprew{\macro{\mdp}}(\lozenge T) \leq \ub.
    \label{eq:lbandub}
\end{align}

\paragraph{Uncertain macro-MDPs.}
We capture the a-priori uncertainty about the subMDP results in an uncertain macro-MDP, a particularly shaped \emph{parametric} MDP.
\begin{definition}[Uncertain macro-MDP]
Let $\mdp$ be a hMDP with $n$ non-trivial $\textbf{S}_i$ partitions and $S_\mdp$ partitioned as $S_\mdp = \bigcup \textbf{S}_i \cup S'$. The \emph{uncertain macro-MDP} is defined as 
$\umacro{\mdp} \colonequals   \langle S' \cup  \{ \entry_i \mid 1 \leq i \leq n \}, \Act_\mdp, \iota_\mdp, \vec{x}, P, r, T_\mdp \rangle$ with parameters $\vec{x} \colonequals \{ p_{i,j}, q_i \mid 1 \leq i \leq n, 1 \leq j \leq Y \}$ where 
$Y = |\succ{\mdp}{\textbf{S}_i}|$. $P$ and $r$ given by
\[ 
    P(s, \act, s') \colonequals \begin{cases} p_{i,j} &\text{if } s \in \textbf{S}_i \text{ and } \\ &  ~s' = \succ{\mdp}{\textbf{S}_i}_j, \\ 
    P_\mdp(s,\act, s') & \text{otherwise,} \end{cases} \quad r(s) \colonequals \begin{cases} q_i &\text{if } s \in \textbf{S}_i, \\ 
    r_\mdp(s) & \text{otherwise.}  \end{cases}
\]
\end{definition}
\begin{remark}
Whenever $\mdp_i$ and $\mdp_{i'}$ are isomorphic, we may reduce the parameters and replace each occurrence of $p_{i',j}$ with $p_{i,j}$ and each occurrence of $q_{i'}$ with $q_i$.
\end{remark}
The uncertain macro-MDP can be instantiated to coincide with the macro-MDP by setting the parameters accordingly. 
\begin{theorem}
\label{thm:uncertainty}
Let $\mdp$ be a hMDP, $\macro{\mdp}$ the associated unique macro-MDP, and $\umacro{\mdp}$ the associated uncertain macro-MDP with parameters $p_{i,j}$ and $q_i$. Let $u^*$ be a parameter valuation with $u^*(p_{i,j})= \res_i(j)$ and $u^*(q_i)= \res_i(Y)$ for all $i,j$. Then:   
\[ 
\umacro{\mdp}[u^*] = \macro{\mdp} \]
\end{theorem}
\paragraph{Proof sketch.}
The construction of the uncertain macro-MDP and the macro-MDP only differs in the assignment of probabilities. We set $u$ here as in the characterisation in~\eqref{eq:mmdpwresults} and thus the equality follows. \qed

\paragraph{Computing bounds.}
Assume for now that we can derive some (trivial) sound bounds on the results vector for any subMDP $\mdp_i$\footnote{We discuss our approach in~Sec.~\ref{sec:templates}, alternatively, one may use bounds  from, e.g., \cite{DBLP:conf/cav/Baier0L0W17}.}. 
\begin{definition}[Sound bounds on results]
\label{def:soundbounds}
For $\mdp_i$, the vectors $\lbres_i$ and $\ubres_i$ are \emph{sound bounds} if the following pointwise inequality holds  \begin{align}
    \lbres_i \leq \res_i \leq \ubres_i.
    \label{eq:properbounds}
\end{align}
\end{definition}
These bounds on properties in the subMDP correspond to bounds on the parameters of the uncertain macro-level MDP $\umacro{\mdp}$. Let us formalize this idea. 
\begin{definition}[Suitable parameter region]
 Given $u^*$ from Theorem~\ref{thm:uncertainty}. The bounds $\LB, \UB$ are \emph{suitable} if $\LB \leq u^*  \leq \UB$. For suitable $\LB, \UB$, the region $[[\LB, \UB]]$ is called \emph{suitable}.
\end{definition}
Using this notion, sound bounds $\lbres_i$ and $\ubres_i$ thus yield suitable bounds $\LB(x), \UB(x)$ for all $x \in \bigcup_j p_{i,j} \cup \{ q_i \}$. Combined, the sound bounds for every $i$ yields a suitable region.
Formally:
\begin{lemma}
\label{lem:suitableboundsfromproperbounds}
Given sound bounds $\lbres_i, \ubres_i$ for each $i$, there exists a trivial mapping $\restoreg$ s.t.\ $\restoreg(\lbres_1, \hdots \lbres_n, \ubres_1, \hdots \ubres_n)$ is a suitable region.
\end{lemma}
With the suitable region we can apply verification on the parametric MDP.
\begin{lemma}
\label{lem:region}
   Let $R$ be a suitable region. Then: \[ \min_{u \in R} \maxexprew{\umacro{\mdp}[u]}(\lozenge T) \leq \maxexprew{\mdp}(\lozenge T) \leq \max_{u \in R} \maxexprew{\umacro{M}[u]}(\lozenge T).  \]
\end{lemma}
\paragraph{Proof sketch.}
We observe that the inequalities follow from the fact that $u^* \in R$ with $u^*$ as in Thm.~\ref{thm:uncertainty}. By that theorem, $\maxexprew{\umacro{\mdp}[u^*]}(\lozenge T) = \maxexprew{\macro{\mdp}}(\lozenge T)$. The statement then follows from Thm.~\ref{thm:macrocorrect}. \qed
 
\noindent  
From the bounds that we can compute using a suitable region, we then set $\lb$ and $\ub$ for Eq.~\eqref{eq:lbandub}:
\begin{align}\label{eq:lbdef} \lb \leq \min_{u \in R} \maxexprew{\umacro{\mdp}[u]}(\lozenge T) \leq \color{black!40} \maxexprew{\mdp}(\lozenge T) \color{black} \leq \max_{u \in R} \maxexprew{\umacro{M}[u]}(\lozenge T) \leq \ub. \end{align}
Computationally, we may use parameter lifting~\cite{DBLP:conf/atva/QuatmannD0JK16} to find these values.

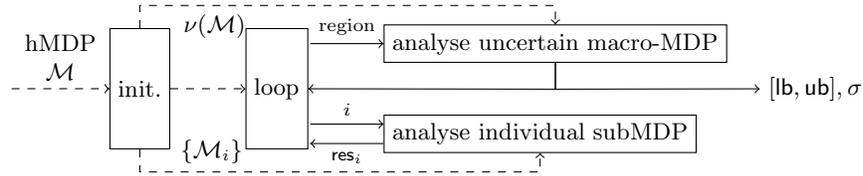
\begin{figure}[t]
    \centering
    \begin{tikzpicture}
     
       \node[rectangle, draw, minimum height=1.6cm] (init) {init.};
       \node[left=1.3cm of init] (in) {};
       \node[rectangle, draw,right=of init, minimum height=1.6cm] (ar) {loop};
       \node[rectangle, draw, right=of ar, yshift=6mm] (umc) {analyse uncertain macro-MDP};
       \node[right=6cm of ar] (result) {$[\lb, \ub], \sched$};
       
       \node[rectangle,draw, right=of ar, yshift=-6mm] (mc) {analyse individual subMDP};

       \draw[->, dashed] (in) -- node[above, align=center] {hMDP\\$\mdp$} (init);
       \draw[->, dashed] (init.north) -- +(0,0.3) -| node[pos=0.09, below] {$\umacro{\mdp}$} (umc.north);
       \draw[->, dashed] (init.south) -- +(0,-0.3) -| node[pos=0.09, above] {$\{ \mdp_i \}$} (mc.south);
       \draw[->, dashed] (init) -- (ar);
       \draw[->] (umc.south) -- +(0,-0.35) -- (ar);
       \draw[->] (umc.south) -- +(0,-0.35) -- (result);
       
       \node[inner sep=0pt] at (ar.east |- umc.west) (arumcanchor) {};
       \draw[->] (arumcanchor) -- node[above] {\scriptsize region} (umc.west);
       
      \node[inner sep=0pt] at (ar.east |- mc.west) (armcanchor) {};
      \node[above=0.8mm of armcanchor,inner sep=0pt] (armcanchorout){};
      \node[below=0.8mm of armcanchor, inner sep=0pt] (armcanchorin) {};
      \draw[->] (armcanchorout) -- node[above] {\scriptsize $i$} (mc.west |- armcanchorout);
      \draw[<-] (armcanchorin) -- node[below] {\scriptsize $\res_i$} (mc.west |- armcanchorin);
    \end{tikzpicture}
    \caption{Analysing hMDPs via uncertain macro-MDPs via  individual refinement.}
    \label{fig:loop1}
\end{figure}

\paragraph{Refinement loop.}
The complete anytime algorithm is summarized in Fig.~\ref{fig:loop1}. We start with an hMDP $\mdp$ and  extract the uncertain macro-MDP $\umacro{\mdp}$ and the subMDPs $\{\mdp_i\}$\footnote{For efficiency, one must implement extraction without first computing an explicit representation of $\mdp$.}. Furthermore we compute (trivial) sound bounds on $\lbres_i \leq \res_i \leq \ubres_i$. This leads to a suitable region $[[\LB, \UB]] = \restoreg(\lbres_1, \ubres_1, \hdots)$. Then,  we may at any time compute the bounds $\lb, \ub$ on the expected reward in the hMDP $\mdp$ by analysing $\umacro{\mdp}$ on the region $[[\LB, \UB]]$. To tighten these bounds, we must first refine the suitable region. Therefore, we  analyse individual subMDPs $\mdp_i$ and compute $\res_i$ and thus $u^*(x)$ for $x \in \cup_j p_{i,j} \cup q_i$. This refines the suitable  bounds such that $\LB(x) = u^*(x) = \UB(x)$ for $x \in \cup_j p_{i,j} \cup q_i$.
We call this refinement \emph{individual refinement}. The new region is suitable and Theorem~\ref{thm:uncertainty} ensures correctness of the refinement. As we only have finitely many subMDPs, we obtain $\lb = \ub$ after finitely many steps.

\begin{mdframed}
\textbf{Anytime version of the enumeration baseline.} Individually refine any subset of subMDPs, then analyse the uncertain macro-MDP $\umacro{\mdp}$.
\end{mdframed}

\subsection{Set-based subMDP analysis}
\label{sec:setbasedanalysis}
Next, we aim to provide an alternative refinement procedure that analyses a set of subMDPs at once, i.e., that refines the suitable bounds for a set of parameters at once. We denote the set of goal states for all subMDPs as $G$\footnote{Formally, we label the goal states and use $G$ to refer to denote those states.}.

\paragraph{Adequate abstractions.}
We aim to compute sound bounds on the results for a set of subMDPs such that the bounds are sound for every individual subMDP in this set. We generalize Def.~\ref{def:soundbounds} as follows:
The (lower and upper) bounds $\lbres_{I}, \ubres_{I}$ are \emph{sound}, if they are sound (lower and upper) bounds for every $\res_i$, $i \in I$. 
\begin{lemma}
\label{lem:soundsetbounds}
Let $\lbres_I$ satisfy the following inequations using $0 \leq j < Y$: \begin{align}\label{eq:lbresdef} \lbres_{I}(Y) \leq \min_{i} \maxexprew{\mdp_i}(\lozenge G)\quad\text{ and }\quad\lbres_I(j) \leq \min_i \min_\sched \prob{\mdp_i[\sched]}(\lozenge G).\end{align}
Then, $\lbres_I$ is a sound lower bound.
\end{lemma}
\noindent\emph{Proof sketch.} We must show $\lbres_I \leq \res_i$ for each $i \in I$. By definition for each $1 \leq j \leq Y$, $\lbres_I(j) \leq \min_{i' \in I} \res_{i'}(j)$ and trivially $\min_{i' \in I} \res_{i'}(j) \leq \res_i(j)$.
\qed

We omit the analogous statement for $\ubres$\footnote{where min becomes max and inequalities flip.}. In Sec.~\ref{sec:templates}, we discuss a particular approach to obtain these bounds, i.e., the right hand sides of the equations in Eq.~\ref{eq:lbresdef}. Here, we update the algorithm sketch to handle this alternative refinement.

\begin{remark}\label{rem:schedulers}
We cannot compute the optimal policy $\sigma_i$ for the subMDP $\mdp_i$ in this setting. Thus, we must compute probability bounds for all policies, which may make these bounds weak. Some optimizations are possible as some actions can in fact be excluded. More importantly, however, is that for cases within Prop.~\ref{prop:sufficient} the policy $\sigma_i$ is irrelevant.
\end{remark}

\paragraph{Updated algorithm.}
We update the loop from Fig.~\ref{fig:loop1}: Rather than refining using a single $i$, we refine using a set $I$. Instead of $\res_i$, we use Lem.~\ref{lem:soundsetbounds} to compute sound bounds $\lbres_I, \ubres_I$ and call this \emph{set-based refinement}.
We may set $\lbres_i = \lbres_I$ for each $i \in I$. Then, we can compute a new suitable region via Lemma~\ref{lem:suitableboundsfromproperbounds}.  
With the suitable region, we can still utilise Eq.~\eqref{eq:lbdef} to compute an approximation $[\lb, \ub]$.
However, for completeness we must ensure that if $|I|=1$, the upper and lower bounds coincide, i.e., $\lbres_{\{i\}} = \ubres_{\{ i\}}$ for every $i$.
That can  be ensured by using individual subMDP refinement when $|I|=1$.

\begin{mdframed}
\textbf{Idea:} We may improve the anytime algorithm by iteratively considering sets of subMDPs and extract sound bounds.
\end{mdframed}
We now first discuss the set-based analysis of multiple subMDPs $\mdp_i$.
We  clarify the realization of the loop box in Sec.~\ref{sec:choices}.

\subsection{Templates for set-based subMDP analysis}
\label{sec:templates}
\label{sec:psubmdps}
We present an instance of set-based subMDP analysis where the subMDPs can be described as instantiations of a parametric MDPs.

\paragraph{Parametric templates.}
We observe that the subMDPs are often similar, e.g., they define sending a file over a channel, exploring a room, in different conditions. We capture this similarity as follows:
Let $\{ \classmdp_1, \hdots \classmdp_m \}$ define a set of parametric MDPs, where we call each pMDP a \emph{template}. 
In particular, for a hierarchical MDP $\mdp$ with partitioning $\textbf{S}_1, \hdots \textbf{S}_n$ and corresponding subMDPs $\mdp_1,\hdots, \mdp_n$
a subMDP $\mdp_i$ is an instantiation of  template $\classmdp_j$ and parameter instantiation $v$ \footnote{We use $v$ instead of $u$ to avoid confusion with the instantiations for pMDP $\umacro{\mdp}$.}, if $\mdp_i = \classmdp_{j}[v]$.
For a concise description, this paper considers hMDPs over a single template $\classmdp$ and, for any $I \subseteq \ntparti$, we denote $V_I \colonequals \{ v_1, \hdots, v_n \}$ the finite (multi)set of parameter instantiations for the pMDP  $\classmdp$ such that $\classmdp[v_i] = \mdp_i$. 

\paragraph{Abstractions from templates.}
In terms of the templates, Lem.~\ref{lem:soundsetbounds} requires us to bound the expected rewards $\maxexprew{\classmdp[v]}(\lozenge G)$ for all $v\in V_I$.
We realize this by defining the smallest region $\toReg(V_I) \supseteq V_I$. 
For this region, we obtain expected rewards by computing the minimum maximal reward in $\toReg(V_I)$. That is: 
\[ \lbres_I(Y) \eqdef \min_{v \in \toReg(V_I)} \maxexprew{\classmdp[v]}(\lozenge G) \quad \leq\quad \min_i \maxexprew{\mdp_i}(\lozenge G).  \]
We handle the probabilities equally while taking into account the quantification over the policies. Following Lem.~\ref{lem:soundsetbounds}, these bounds are sound. Upper bounds are handled analogously. Computationally, we again use parameter lifting~\cite{DBLP:conf/atva/QuatmannD0JK16} to find these bounds. 
We can easily refine:
Whenever we split $I$ (or equally, $V_I$), we can compute (potentially) smaller regions $\toReg(V_I)$.

 \begin{figure}[t]
     \centering
    \begin{tikzpicture}
     
       \node[rectangle, draw, minimum height=1.6cm] (init) {init.};
       \node[left=1.3cm of init] (in) {};
       \node[rectangle, draw,right=of init, minimum height=1.6cm] (ar) {loop};
       \node[rectangle, draw, right=of ar, yshift=6mm] (umc) {analyse uncertain macro-MDP};
       \node[right=6cm of ar] (result) {$[\lb, \ub], \sched$};
       
       \node[rectangle,draw, right=of ar, yshift=-6mm] (mc) {analyse set of subMDPs};

       \draw[->, dashed] (in) -- node[above, align=center] {hMDP\\$\mdp$} (init);
       \draw[->, dashed] (init.north) -- +(0,0.3) -| node[pos=0.09, below] {$\umacro{\mdp}$} (umc.north);
       \draw[->, dashed] (init.south) -- +(0,-0.3) -| node[pos=0.09, above] {{\color{red} $\classmdp$}} (mc.south);
       \draw[->, dashed] (init) -- (ar);
       \draw[->] (umc.south) -- +(0,-0.35) -- (ar);
       \draw[->] (umc.south) -- +(0,-0.35) -- (result);
       
       \node[inner sep=0pt] at (ar.east |- umc.west) (arumcanchor) {};
       \draw[->] (arumcanchor) -- node[above] {\scriptsize region} (umc.west);
       
      \node[inner sep=0pt] at (ar.east |- mc.west) (armcanchor) {};
      \node[above=0.8mm of armcanchor,inner sep=0pt] (armcanchorout){};
      \node[below=0.8mm of armcanchor, inner sep=0pt] (armcanchorin) {};
      \draw[->] (armcanchorout) -- node[above] {\scriptsize $I$} (mc.west |- armcanchorout);
      \draw[<-] (armcanchorin) -- node[below] {\scriptsize bounds} (mc.west |- armcanchorin);
    \end{tikzpicture}
    \caption{Analysing hMDPs with set-based refinement on templated subMDPs.}
    \label{fig:loop2} 
 \end{figure}
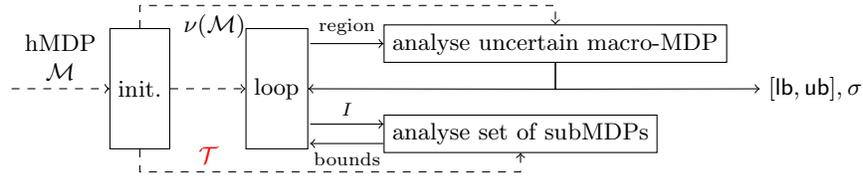
In Fig.~\ref{fig:loop2}, we depict our method. In contrast to Fig.~\ref{fig:loop1}, we pass the template $\classmdp$ rather than the individual subMDPs.
Furthermore,  we now compute initial sound bounds via the analysis of  the template (i.e., of $V_I$) and must pass the mapping from $I$ to $V_I$ to clarify the shape of the subMDPs.
\begin{mdframed}
\textbf{Abstraction-Refinement} on the subMDPs provides increasingly tight suitable regions for  the uncertain macro-MDP from the anytime baseline.
\end{mdframed}



\section{Implementing the Abstraction-Refinement Loop}
\label{sec:choices}
\label{sec:alg}
Alg.~\ref{alg:main} outlines a basic implementation of the idea sketched in Fig.~\ref{fig:loop2}. We detail this implementation and then discuss an essential improvement.

\begin{algorithm}[t]
\begin{algorithmic}[1]
\State Construct macro-MDP $\umacro{\mdp}$, class-MDP $\classmdp$, and $V_\ntparti$ from high-level description. 
\State $Q \gets \{ \langle I = \ntparti, \text{bounds}=[0, \infty), \text{weightedvals}=\ntparti \rightarrow \{1\} \rangle  \}$
\State $\lb \gets 0$; $\ub \gets \infty$; $\#\text{iter} = 0$; $\text{Res} \gets \emptyset$
\While{$\eta \cdot \ub > \lb$}
    \State $R \gets Q.pop()$ \label{algline:startloop}   \Comment{Use priority}
    \If{$R.I = \{ i \}$}
        \State $\text{Res}[i] \gets check\_one(\classmdp[v_i])$ \Comment{Computes $\res_i$}
    \Else
    \State $R.bounds \gets check\_set(\classmdp, \toReg(V_{R.I}))$ \Comment{Computes $\lbres_{R.I},\ubres_{R.I}$}
    \State $Q \gets Q \cup split(R)$ \Comment{Split $R.I$, keep bounds and weights}
    \EndIf \label{algline:endresref}
    \If{$\#\text{iter}~\textsl{mod}~8 = 1$ or $Q$ is empty} \label{algline:mcstart}
        \State $R' \gets \restoreg(extract(Q, \text{Res}))$  \Comment{Compute suitable region via Lem~\ref{lem:suitableboundsfromproperbounds}}\label{line:main:mcmacromdp}
        \State $\lb, \ub \gets check\_set(\umacro{\mdp}, R')$
    \EndIf\label{algline:mcend}
\EndWhile
\end{algorithmic}
\caption{Algorithm for Abstraction-Refinement Procedure}
\label{alg:main}
\end{algorithm}

 We construct $\umacro{\mdp}$, $\classmdp$, and (the implicit) mapping $V \colon \ntparti \rightarrow V_\ntparti$ to map subMDPs to instantiations of $\classmdp$ from a suitable high-level representation. 
 We initialize a priority queue with triples  that represent sets of template instantiations: $I$ such that  $V_I \colonequals \{ v_i \colonequals V(i) \mid i \in I\}$ contains all valuations $v$ such that $\classmdp[v]$ is a subMDP of $\mdp$. We initially store bounds reflecting $\lbres_I$ and $\ubres_I$ as well as weights for the computation of the priority (see below). Initially, we assume that $\lb =0$ and $\ub = \infty$, we count the number of iterations in $\#\text{iter}$. $\text{Res}$ is map for storing result vectors.
 The algorithm now refines $\lb$ and $\ub$ until the gap between $\lb$ and $\ub$ is sufficiently small. 

The main loop now iteratively refines $\lb,\ub$ by first refining $\lbres_I$ and $\ubres_I$, by splitting $I$ and model checking $\classmdp$ w.r.t.\  subsequently smaller regions $\toReg(V_I)$ (l.~\ref{algline:startloop}-\ref{algline:endresref}): Therefore, we take a set $R$ from the queue. If $R.I =\{ i \}$ is a singleton, we compute $\lbres_{R.I} = \res_{i} = \ubres_{R.I}$ and store this result. Otherwise, we apply model checking to the pMDP $\classmdp$ w.r.t.\ the region representation of $R.I$. We then split $R.I$, by splitting $I$ into (here) two subsets. For splitting $I$, we use the geometric interpretation of $\toReg(V_I)$ as a subset of $\mathbb{R}^{|\vec{y}|}$, where we then split along one of the axis into two equally large subsets. 
Every $k$ (we use $k=8$) iterations, we analyse the macro-MDP (l.~\ref{algline:mcstart}-\ref{algline:mcend}). From $Q$ and $\text{Res}$ we extract the proper bounds $\lbres_i, \ubres_i$ from $Res[i]$ if possible and from $Q$ using $R.\text{bounds}$ for $R$ such that $i \in R.I$ otherwise. Then via $\restoreg(\lbres_1, \ubres_1,\hdots)$ from  Lem.~\ref{lem:suitableboundsfromproperbounds} we compute a suitable region $R'$. We analyse the uncertain macro-MDP to obtain $\lb$ and $\ub$ in accordance with  Eq.~\eqref{eq:lbdef}.

Finally, we discuss the priority function: If we a-priori naively assume that each subMDP contributes an equal amount to the overal minimal expected reward in the hMDP (weights are all one) then the following priority function: $|R.\text{bounds}| \cdot \sum_{v \in I} \text{R.weights}(v)$ computes priorities that correlate with how much computing $\res_i$ for all $i \in I$ would reduce the gap between $\lb$ and $\ub$.


\paragraph{Termination and correctness argument.}
Algorithm~\ref{alg:main} terminates. We split in such way that $\max_{I \in Q} |I|$ monotonically decreases. Thus, eventually $Q$ is empty and $\text{Res}$ contains results for all subMDPs. Then, $R'$ is a point region and checking $\umacro{\mdp}$ with this point region ensures that $\lb = \ub$. Correctness follows as $R'$ is always suitable, see Eq.~\eqref{eq:lbdef}.

\paragraph{Computing expected visits.}
Based on our empirical evaluation we added one crucial improvement: While the algorithm above assumed that all subMDPs (or states in the macro-MDP) are equally important, that assumption is generally inadequate. 
Roughly, only states reached by the optimal policy contribute at all (provided the bounds are tight enough that we can identify these states). The reachable states are  weighted by the expected number of visits of these states. We compute an approximation of this expected number of visit by computing the currently optimizing policy (a by-product of l.~\ref{line:main:mcmacromdp})  and compute the center of $R'$; this results in a MC for which we can compute the number of expected visits by a standard equation system~\cite{Put95}. Additionally, we  update the weights for the regions in the queue based on these new results. We remark that this also makes the priority function more useful. 

\paragraph{Interleaving individual refinement.}
Furthermore, for a subMDPs for which the expected number of visits is large\footnote{In our implementation, we define this as subMDPs where the expected number of visits is in the top $1 + \nicefrac{1}{16} \cdot \#\text{iter}$ percent, but not more than 150 at a time.} are individually analysed (and the points are removed from the region in the queue). This optimization reduces the need to split the corresponding regions until we obtain tight bounds.

\section{Experiments}

\paragraph{Implementation.}
We implemented \textsc{level-up}\footnote{The source code and executables, the benchmarks, logfiles and utilities are all available in an archived Docker container: \url{https://doi.org/10.5281/zenodo.6524787} }, a prototype on top of the python bindings for Storm~\cite{DBLP:journals/corr/abs-2002-07080}. \textsc{level-up} analyses hierarchical MDPs by taking two MDPs, each provided as probabilistic program descriptions in the PRISM format:
One MDP that encodes the (uncertain) macro-MDP and one that describes the parametric template for the subMDPs. 
The parameter instance of the subMDP can be deduced as a function of the high-level variable assignment of the macro-MDP states. 
For technical reasons, the prototype currently provides support for subMDPs with one or two successor states -- arguably the setting in which we expect our prototype to perform best. For subMDPs with a single successor state, the uncertain macro-MDP may be represented as an (parameter-free) MDP with interval-valued rewards.
For two successors, we include support of the extension of Sec.~\ref{sec:localcase} where the successor aims to optimize reaching a fixed successor state.

\paragraph{Setup.} 
We investigate the scalability and the quality of the approximation over time. Therefore, we run our prototype on an MacBook 2020 M1 with an 8 GB RAM limit.
We compare the enumerative baseline from Sec.~\ref{sec:macromdp} with Alg.~\ref{alg:main}. Both exploit the hierarchical nature of the MDP.  We qualitatively compare to standard model checking on the flat MDP, see below.   We use a collection of benchmarks reflecting networks, job schedulers and robots. 

\begin{table}[t]
\centering
\caption{Benchmark statistics, runtimes of the approaches, and details for Alg.~\ref{alg:main}.}
\scalebox{0.82}{
\begin{tabular}{lcrrrrrrr||r|rrr||rrrrr}
Name & Inst & $|S_\mdp|$ & $|\ntparti|$ &
$|S_{\macro{\mdp}}|$ & $|\text{A}_{\macro{\mdp}}|$   &
$|S_\classmdp|$ & $|\text{A}_\classmdp|$ & 
$t_\text{init}$ & $t_\text{enum}$ & $t_{50}$ & $t_{90}$ & $t_{95}$ & iter. & indrf. & $\stackrel{\text{um}}{\%}$ & $\stackrel{\text{sr}}{\%}$  & $\stackrel{\text{ir}}{\%}$  \\\hline
\textsf{corr} &	\tiny{11,10,50} &	$10^\textbf{7}$ &	$624$ &	$255576$ &	$541704$ &	$15000$ &	$65006$ &	$<1$ &	$16$ &	$3$ &	$9$ &	$\mathbf{13}$ &	$17$ &	$14$ &	$2$ &	$67$ &	$2$\\
\textsf{corr} &	\tiny{11,8,100} &	$10^\textbf{8}$ &	$624$ &	$254376$ &	$539040$ &	$60000$ &	$260006$ &	$<1$ &	$100$ &	$10$ &	$45$ &	$\mathbf{45}$ &	$9$ &	$16$ &	$2$ &	$80$ &	$4$\\
\textsf{corr} &	\tiny{11,8,200} &	$10^\textbf{8}$ &	$624$ &	$254376$ &	$539040$ &	$240000$ &	$1040006$ &	$2$ &	$689$ &	$51$ &	$313$ &	$\mathbf{568}$ &	$17$ &	$30$ &	$0$ &	$92$ &	$4$\\
\textsf{corr} &	\tiny{13,11,50} &	$10^\textbf{7}$ &	$768$ &	$1024344$ &	$2172432$ &	$15000$ &	$65006$ &	$3$ &	$21$ &	$8$ &	$\mathbf{18}$ &	$25$ &	$17$ &	$17$ &	$5$ &	$36$ &	$1$\\
\textsf{corr1} &	\tiny{17,14,75} &	$10^\textbf{8}$ &	$1056$ &	$34200$ &	$83160$ &	$33750$ &	$146256$ &	$<1$ &	$90$ &	$4$ &	$21$ &	$\mathbf{38}$ &	$17$ &	$43$ &	$0$ &	$84$ &	$8$\\
\textsf{corr1} &	\tiny{18,15,75} &	$10^\textbf{8}$ &	$1128$ &	$39576$ &	$96768$ &	$33750$ &	$146256$ &	$<1$ &	$98$ &	$4$ &	$38$ &	$\mathbf{38}$ &	$17$ &	$45$ &	$0$ &	$84$ &	$8$\\
\textsf{corr1} &	\tiny{25,20,75} &	$10^\textbf{8}$ &	$1632$ &	$89136$ &	$224160$ &	$33750$ &	$146256$ &	$<1$ &	$168$ &	$5$ &	$44$ &	$\mathbf{67}$ &	$25$ &	$102$ &	$1$ &	$80$ &	$14$\\
\textsf{mail} &	\tiny{10} &	$10^\textbf{9}$ &	$173857$ &	$793971$ &	$1088152$ &	$2801$ &	$3601$ &	$4$ &	$552$ &	$8$ &	$21$ &	$\mathbf{48}$ &	$57$ &	$658$ &	$29$ &	$2$ &	$4$\\
\textsf{mail} &	\tiny{12} &	$10^\textbf{9}$ &	$236802$ &	$1446551$ &	$2023504$ &	$2801$ &	$3601$ &	$8$ &	$738$ &	$16$ &	$43$ &	$\mathbf{130}$ &	$97$ &	$703$ &	$42$ &	$1$ &	$2$\\
\textsf{netw} &	\tiny{30,50} &	$10^\textbf{8}$ &	$9801$ &	$437823$ &	$437823$ &	$4026$ &	$4026$ &	$1$ &	$23$ &	$\mathbf{8}$ &	$33$ &	$46$ &	$217$ &	$150$ &	$60$ &	$1$ &	$1$\\
\textsf{netw} &	\tiny{30,80} &	$10^\textbf{8}$ &	$9801$ &	$437823$ &	$437823$ &	$10041$ &	$10041$ &	$1$ &	$62$ &	$8$ &	$34$ &	$\mathbf{48}$ &	$217$ &	$150$ &	$59$ &	$3$ &	$3$\\
\textsf{netw} &	\tiny{50,80} &	$10^\textbf{8}$ &	$9801$ &	$1025883$ &	$1025883$ &	$10041$ &	$10041$ &	$2$ &	$62$ &	$\mathbf{16}$ &	$94$ &	$112$ &	$225$ &	$150$ &	$62$ &	$1$ &	$1$\\
\textsf{sdn} &	\tiny{5,12,4,4} &	$10^\textbf{8}$ &	$23375$ &	$128386$ &	$128386$ &	$13506$ &	$16855$ &	$<1$ &	$62$ &	$2$ &	$\mathbf{20}$ &	$112$ &	$289$ &	$305$ &	$2$ &	$17$ &	$11$\\
\textsf{sdn} &	\tiny{5,8,4,4} &	$10^\textbf{8}$ &	$23375$ &	$128386$ &	$128386$ &	$2802$ &	$3455$ &	$<1$ &	$98$ &	$1$ &	$5$ &	$\mathbf{15}$ &	$281$ &	$305$ &	$13$ &	$17$ &	$8$\\
\textsf{sdn} &	\tiny{6,8,4,4} &	$10^\textbf{9}$ &	$126337$ &	$408227$ &	$408227$ &	$2802$ &	$3455$ &	$2$ &	$519$ &	$5$ &	$46$ &	$\mathbf{394}$ &	$3057$ &	$305$ &	$27$ &	$7$ &	$0$
\end{tabular}
}
\label{tab:benchmarkstats}
\end{table}

\paragraph{Results.}
We consider instances that we summarize in Tab.~\ref{tab:benchmarkstats}. In particular, we give the benchmark name and instance for reference, the approximate number of states in the hierarchical MDP (computed from the macro-MDP and the subMDPs), the number of nontrivial partitions, and the number of states and actions in the (uncertain) macro-MDP and subMDPs, respectively. Then, we give the time to setup the data structures from the high-level representation $t_\text{init}$ in seconds. We highlight that a flat representation of all our benchmarks has at least $10^7$, often more, states.
As a reference, we present the performance of the enumerative baseline from Sec.~\ref{sec:macromdp}. The performance of this approach is positive as it enables the verification of huge MDPs. A TO indicates ${>}1200$ seconds.
To scale to either larger subMDPs or more subMDPs, we use the abstraction-refinement loop. To reflect the anytime nature, we list three run times for terminating when $\eta \cdot \ub \leq \lb$ with $\eta \in \{ 0.5, 0.9, 0.95 \}$ respectively. The largest time faster than the enumerative baseline is highlighted (further to the right is better for the abstraction-refinement). For $\eta=0.95$, we give details: The number of iterations (iter), the number of individual refinements based on the improvement from Sec.~\ref{sec:alg}, and the fraction of time spent on model checking the uncertain macro-MDPs $\%_\text{um}$, the set-refinements $\%_\text{sr}$, and the individual refinements $\%_\text{ir}$, respectively.


    

\paragraph{Discussion.}
Before we discuss details of the results, let us clarify that \emph{exploiting the hierarchical structure is essential}. MDPs with ${\approx}10^8$ are at the limit of what fits in around  8GB of memory \footnote{Assuming 128 byte per state, i.e., 8 doubles and 16 (32-bit) ints, as used in Storm.}. Symbolic methods based on MTBDDs easily scale beyond these sizes, but ---noting that the subMDPs are all slightly different--- the models we consider lack the necessary symmetry that make MTBDDs compact. Thus, support for hierarchical MDPs is a necessary step forward.

Regarding the abstraction-refinement: While a larger study may be necessary, we can start with two standard observations: The abstraction-refinement loop is significantly faster on $\eta \leq 0.9$. As $\eta \rightarrow 1$, coarse abstractions are insufficient. Furthermore, the efficiency of the abstraction-refinement heavily depends on the particular structure.
That being said, the approach outperforms the enumerative approach, especially for $\eta = 0.9$, and up to more than an order of magnitude.
This happens even if $\ntparti$ is rather small, or if, e.g., $\classmdp$ is small. We furthermore observe that for large $\ntparti$, the bookkeeping in python becomes a bottleneck.  We think these observations are promising: we left many options for further optimizations and tweaking towards particular examples on the table.
However, for models where most time is spent on model checking the macro-level MDP, the approach is less suitable.
We furthermore conjecture that tailored algorithms may exploit some of these dimensions, e.g., when there is the macro-MDP or the subMDPs are indeed MCs or perhaps acyclic, depending on the number of parameters and their influence~\cite{DBLP:conf/tacas/SpelJK21}, or based on the relative weight of the uncertain rewards compared to rewards in the macro-MDP. 


\section{Related work}
\label{sec:related}

In the model-free reinforcement learning (RL) setting, hierarchical models are popular. An excellent, recent survey is given in \cite{DBLP:journals/csur/PateriaSTQ21}.
Our work generalizes the solution techniques on hierarchical MDPs that assume that these subMDPs are the same. In RL, this assumption is treated liberally, and the methods provide only weak error bounds. In contrast, our model-based approach provides error-bounds in every step, and the error disappears in finitely many steps.

Hierarchical abstractions are used to analyse large MDPs in~\cite{DBLP:conf/ijcai/BarryKL11}. There, the goal is to find a policy that almost optimizes the reward. Rather than preimposing a hierarchy, the algorithm aims to find a hierarchy and define the goal states of the subMDP such that the model admits local policies. Instead, our solution can find the optimal policy and in particular gives strict error bounds at the cost of requiring a high-level model that induces the hierarchy. 
An symbolic approach for continuous MDP, where the transition probabilities are the result of an associated LP, has recently been discussed in \cite{DBLP:conf/ijcai/JeongJS21}. 
 An hierarchical SCC-decomposition~\cite{DBLP:conf/qest/AbrahamJWKB10} aims to accelerate the process of solving a (given, monolithic) Markov chain.  
 The computation of reward-bounded properties~\cite{DBLP:journals/jar/HartmannsJKQ20} generalizes topological value iteration and their notion of episodes mildly resembles an hierarchical approach but no uncertainty is assumed or used in the approach.
 The probabilistic model checker PAT~\cite{DBLP:conf/cav/SongSLD12} analyses a hierarchical probabilistic timed automaton given as a process algebra. The hierarchy is not exploited in the solving process.


While symbolic approaches, often on decision diagrams, exploit the transition system by compressing the data structures, abstractions aim to yield smaller systems that may assess an approximation for the sought-for values. Abstraction-refinement without an imposed hierarchy is explored in~\cite{DBLP:conf/cav/HermannsWZ08,DBLP:journals/fmsd/KattenbeltKNP10,DBLP:conf/tacas/HahnHWZ10}: Refinement amounts to considering a better approximation of the state space.
 In contrast, we impose the hierarchy, the abstraction amounts to an imprecise analysis of this fixed state space and we refine by analysing the state space more precisely (by means of analysing subMDPs at a greater level of detail). Contract-based abstractions (in probabilistic systems) are used to decompose the analysis of systems given by parallel running subsystems~\cite{DBLP:conf/atva/XuGG10,DBLP:conf/atva/FengHKP11,DBLP:conf/tacas/KwiatkowskaNPQ10}. Partial exploration and bounded model checking approaches focus on the most critical paths, i.e., the paths where most of the probability mass lies~\cite{DBLP:conf/atva/BrazdilCCFKKPU14,DBLP:conf/atva/0001DKKW16,DBLP:journals/lmcs/KretinskyM20}, but these approaches do generally not exploit the hierarchical and repetitive structure. The observation that many parts of the system are not critical allows us to weigh the potential benefit of refining the intervals in various parts of the macro-MDP. 
 
Parametric MDPs are commonly used to model and analyse the effects of uncertainty in the precise transitions~\cite{DBLP:conf/cav/HahnHWZ10,DBLP:conf/cav/PuggelliLSS13,DBLP:conf/atva/0001DKKW16}. The methods presented in \cite{DBLP:conf/qest/JansenCVWAKB14,DBLP:conf/icse/FangCGA21} exploit a repetitive structure in parametric MCs to accelerate the construction of closed form solutions and are not applicable to MDPs. 
Parametric models have been used to support the design of systems~\cite{DBLP:journals/jss/CalinescuCGKP18,DBLP:conf/tacas/Andriushchenko021} or their adaption~\cite{DBLP:conf/tacas/BartocciGKRS11,DBLP:conf/tase/ChenHHKQ013}, to find policies for partially observable systems~\cite{DBLP:conf/aaai/Cubuktepe0JMST21}, to analyse Bayesian networks~\cite{DBLP:conf/ecsqaru/SalmaniK21}, and to speed up the analysis of, e.g., software product lines~\cite{DBLP:journals/fac/ChrszonDKB18,DBLP:journals/sttt/BeekL19}. On top of technical differences, none of these approaches uses a hierarchical decomposition of an MDP or uses the results of the analysis in the analysis of a larger MDP. 

\section{Conclusion}
This paper presents a first verification approach that exploits the hierarchical structure natural in many models to accelerate analysing the underlying MDP. An essential ingredient is to separate the two levels in the hierarchy. Then, when analysing the (toplevel) macro-MDP, we may consider subMDPs that have not yet been analysed as epistemic uncertainty. Analysis techniques for uncertain (more precise: parametric) MDPs then enable an online approximation loop that incrementally removes uncertainty in a targeted fashion by analysing more and more subMDPs (more) precisely. 
Three clear directions for future work are to (i)~consider an approach where one lifts the restrictions to locally-optimal policies, (ii)~investigate the applicability to a richer set of temporal properties and (iii)~to allow automatic detection of partitions in, e.g., the Prism language. 


\bibliographystyle{plain}
\bibliography{bibliography.bib}

\clearpage
\pagebreak
\appendix

\section{Proof of Thm.~\ref{thm:macrocorrect}}
\label{app:thmmacrocorrect}

We sketch the main steps. 
Consider the (unique) policies $\sched_i$ from Def.~\ref{def:mmdp} such that $\sched_i$ is optimal for $\mdp_i$. Consider $\parsched \colonequals \bigsqcup_\ntparti \sched_i$. As we assume that $\mdp$ has locally optimal policies, it suffices to show that 
$\maxexprew{\macro{\mdp}}(\lozenge T) = \maxexprew{\mdp[\parsched]}(\lozenge T)$.
Therefore, we take an optimal policy $\sched'$ for the MDP $\mdp[\parsched]$ and show that \[ \maxexprew{\mdp[\parsched]}(\lozenge T) \stackrel{(a)}{=} \exprew{\mdp[\parsched][\sched']}(\lozenge T) \stackrel{(b)}{=} \exprew{\macro{\mdp}[\sched']}(\lozenge T) \stackrel{(c)}{=} \maxexprew{\macro{\mdp}}(\lozenge T). \]
We go through from left to right.
\begin{itemize}\item
(a) holds as $\sched'$ is optimal.
\item For (b), consider that for each state $s = \entry_i, i \in \ntparti$, we have that \[ \Pr_{\mdp[\parsched][\sched']}(s \rightarrow \lozenge \{s'\}) = \Pr_{\macro{\mdp}[\sched']}(s \rightarrow \lozenge \{s'\})\] by construction. Then, for \emph{every} state $s \in S_\macro{\mdp}$ it follows that \[ \Pr_{\mdp[\parsched][\sched']}(s \rightarrow \lozenge T) = \Pr_{\macro{\mdp}[\sched']}(s \rightarrow \lozenge T).\]

Following the characteristic Bellman equations, we furthermore have for every entry state $\entry_i$ -- using that $\parsched = \bigsqcup_\ntparti \sched_i$ and that in $\mdp_i$, reaching successor states is mutually exclusive -- the following holds:
\begin{align*} & \exprew{\mdp[\parsched][\sched']}(s \rightarrow \lozenge T) = \\&  \sum_{s' \in \succ{s}{\textbf{S}_i}} \Pr_{\mdp_i[\sched_i]}(\lozenge \{ s' \}) \cdot \left( \exprew{\mdp[\parsched][\sched']}(s' \rightarrow \lozenge T) + \exprew{\mdp_i[\sched_i]}(s' \rightarrow \lozenge \{ s' \}) \right). 
\end{align*} 
Which we can rearrange to 
\begin{align*} & \exprew{\mdp[\parsched][\sched']}(s \rightarrow \lozenge T)  = \\ &\quad \sum_{s' \in \succ{s}{\textbf{S}_i}} \pr_{\mdp_i[\sched_i]}(\lozenge \{ s' \}) \cdot \exprew{\mdp[\parsched][\sched']}(s' \rightarrow \lozenge T) + \\ & \quad \sum_{s' \in \succ{s}{\textbf{S}_i}} \pr_{\mdp_i[\sched_i]}(\lozenge \{ s' \}) \cdot \exprew{\mdp_i[\sched_i]}(s' \rightarrow \lozenge \{ s' \}) \end{align*}
Or using the structure of the subMDPs (Def.~\ref{def:submdp}):
\begin{align*} & \exprew{\mdp[\parsched][\sched']}(s \rightarrow \lozenge T)  = \exprew{\mdp_i[\sched_i]}(s' \rightarrow \lozenge \{ \bot \}) + \\ &\quad \sum_{s' \in \succ{s}{\textbf{S}_i}} \pr_{\mdp_i[\sched_i]}(\lozenge \{ s' \}) \cdot \exprew{\mdp[\parsched][\sched']}(s' \rightarrow \lozenge T) \end{align*}
Which is in familiar form as characteristic equation system in $\macro{\mdp}$. By substitution we can now conclude that for every state $s \in \macro{\mdp}$ it holds that :
\[ \exprew{\mdp[\parsched][\sched']}(s \rightarrow \lozenge T) = \exprew{\macro{\mdp}[\sched']}(s \rightarrow \lozenge T).\]
    \item 
Regarding (c): From (b) above we know that the expected rewards coincide for every state. By construction of $\sched'$ is optimal on $\mdp[\parsched]$ and these values are  maximal. Now  assume that $\sched'$ is not optimal on $\macro{\mdp}$. From the Bellman characterisation, we know that there exists a state $s$ such that $\sched'(s)$ is not optimal, i.e., changing actions would improve the expected reward from that state onwards. But then $\sched'$ would also not be optimal on $\mdp[\parsched]$. 
\end{itemize}
\qed

\end{document}